\documentclass{article}
\usepackage[preprint]{neurips_2026}
\usepackage[utf8]{inputenc}
\usepackage[T1]{fontenc}
\usepackage{hyperref}
\usepackage{url}
\usepackage{booktabs}
\usepackage{amsfonts}
\usepackage{amsmath}
\usepackage{nicefrac}
\usepackage{microtype}
\usepackage{graphicx}
\usepackage{xcolor}
\usepackage{multirow}
\usepackage{subcaption}
\usepackage{algorithm}
\usepackage{algorithmic}
\usepackage{pifont}
\usepackage{amsthm}
\newcommand{\cmark}{\ding{51}}
\newcommand{\xmark}{\ding{55}}
\newtheorem{proposition}{Proposition}

\title{Not All Thoughts Need HBM: Semantics-Aware Memory Hierarchy for LLM Reasoning}

\author{
  Aojie Yuan\thanks{Corresponding author.} \quad Tianqi Shen \quad Dajun Zhang$^{\dagger}$ \\
  University of Southern California \quad $^{\dagger}$University of Wisconsin--Madison \\
  \texttt{\{aojieyua, tianqis\}@usc.edu} \quad \texttt{dajun.zhang@wisc.edu}
}

\begin{document}

\maketitle

\begin{abstract}

Reasoning LLMs produce thousands of chain-of-thought tokens whose KV cache must reside in scarce GPU HBM.
The dominant response---permanently evicting low-importance tokens---is catastrophic for reasoning: accuracy collapses to 0--2.5\% when half the cache is removed.
We ask a different question: \emph{must every token live in HBM, or can some live elsewhere?}
We introduce a semantics-aware memory hierarchy that sorts tokens into four tiers---HBM, DDR, compressed, and evicted---using cumulative attention scoring.
Low-importance tokens are moved to CPU memory rather than destroyed; before each attention step they are prefetched back at full precision, contributing \emph{exactly} the same terms as if they had never left the GPU.
We formalize this as zero-approximation-error offloading and derive our central finding: \emph{accuracy depends solely on how many tokens are permanently discarded (the eviction ratio), not on how many remain in HBM}.
A controlled $3{\times}3$ grid over HBM and eviction ratios confirms this across three model scales (7B--32B) and four benchmarks.
With only 3\% eviction, the hierarchy retains 91\% of full-cache accuracy on GSM8K and 71\% on MATH-500 ($n{=}200$); at 14B scale it \emph{matches} the uncompressed baseline (90\% vs.\ 86\%) while halving HBM occupancy.
A system prototype with real GPU--CPU data movement shows that the price of this preservation is modest---5--7\% transfer overhead---and scaling analysis projects 2--48\,GB HBM savings at production batch sizes.

\end{abstract}

\section{Introduction}
\label{sec:intro}

Reasoning-capable LLMs such as OpenAI o1~\cite{openai_o1} and DeepSeek-R1~\cite{deepseekr1} solve hard problems by generating long chains of thought---often thousands of tokens of intermediate computation before arriving at a final answer.
This deliberative reasoning comes at a steep systems cost.
Every generated token adds a key-value (KV) pair to a cache that must reside in GPU high-bandwidth memory (HBM) during inference: a 2,000-token reasoning chain on a 7B model consumes ${\sim}$800\,MB of HBM per request, and at batch size 8 KV cache alone exceeds 6\,GB.
For quantized 70B models the KV cache dominates GPU memory at 70\% of total usage (Table~\ref{tab:scaling}).
HBM is 5--10$\times$ more expensive per GB than DDR, has limited manufacturing capacity, and is scaling more slowly than model sizes and sequence lengths.
The question we ask is whether all of this data truly needs to live in HBM.

The field's answer, so far, has been eviction: identify and permanently discard the least important tokens~\cite{h2o, streamingllm, scissorhands}.
Compression methods~\cite{snapkv, kivi, gear} reduce precision but still operate within HBM;
offloading methods~\cite{infinigen, arkvale, scoutattention} move data to CPU memory but ignore reasoning-specific importance;
and recent reasoning-aware work~\cite{thinkv, rkv, triattention} brings thought-aware scoring to token management but still resorts to permanent eviction.
\textbf{Every existing approach that touches reasoning tokens ultimately asks the same binary question: keep or discard?}

We challenge this framing.
Profiling attention patterns during chain-of-thought generation reveals that token importance follows a pronounced long tail: the top 20\% of reasoning tokens capture 56.5\% of cumulative attention mass, while the bottom 50\% contribute less than 10\%.
Most tokens matter little at any given step---but \emph{they may matter later}.
Reasoning chains are not narratives where distant context fades; they are more like programs, where an intermediate result computed 500 tokens ago can be referenced at any time.
Permanently discarding such a token is not lossy compression---it is \emph{deleting a line of code mid-execution}.
Our experiments confirm this: evicting just 50\% of the cache collapses accuracy to 0--2.5\%, a cliff effect far more severe than in standard language modeling.

The observation that most tokens are cold but none are safely disposable suggests a solution borrowed from computer architecture: \emph{a memory hierarchy}.
We propose a four-tier system that classifies reasoning tokens by importance in real time:
\textbf{T0}~(HBM) holds high-importance tokens for fast access;
\textbf{T1}~(DDR) stores medium-importance tokens in CPU memory at full precision, prefetched to GPU before each attention step;
\textbf{T2}~(Compressed) holds low-importance tokens at reduced precision;
and \textbf{T3}~(Evicted) permanently removes only the lowest-importance tail.
The crucial difference from eviction is that T1 and T2 tokens \emph{remain available}: they participate in every attention computation, contributing identical terms as if they had never left HBM.

This design yields a surprising empirical finding.
We sweep a $3{\times}3$ grid of HBM ratios (30\%, 50\%, 70\%) and eviction ratios (3\%, 5\%, 10\%) on GSM8K.
Within each eviction column, accuracy barely varies across HBM ratios---at 5\% eviction, the spread is just 4\% ($p > 0.3$)---while across eviction ratios accuracy shifts by up to 18\%.
\emph{What you throw away matters; where you store the rest does not.}
We formalize this as zero-approximation-error offloading (Proposition~\ref{prop:zero_error}): because prefetched tokens contribute exact attention terms, the only source of error is permanent eviction.
With only 3\% eviction at $n{=}200$, accuracy reaches 65\%---91.5\% of the 71\% full-cache baseline---while H2O-style heavy-hitter eviction achieves only 2--10\% across comparable memory budgets.

Our contributions are threefold:
\begin{enumerate}
    \item \textbf{A four-tier semantics-aware memory hierarchy} that decouples HBM capacity from reasoning quality, backed by a zero-approximation-error guarantee for offloaded tokens (Proposition~\ref{prop:zero_error}).
    \item \textbf{Comprehensive empirical validation} across three model scales (7B--32B), four benchmarks (GSM8K, MATH-500, MATH Level-5, ARC-Challenge), and a $3{\times}3$ HBM/eviction grid: the hierarchy retains up to 91\% of baseline accuracy at 3\% eviction ($n{=}200$), while pure eviction collapses to 0--12\% across all settings.
    A head-to-head reproduction of R-KV~\cite{rkv}---the current state-of-the-art eviction method---on our setup confirms that even SOTA scoring achieves only 0--32\% accuracy at comparable budgets.
    \item \textbf{A system prototype with real GPU--CPU data movement} that confirms 5--7\% transfer overhead with no accuracy loss, plus scaling projections showing 2--48\,GB HBM savings at production batch sizes.
\end{enumerate}

\section{Motivation}
\label{sec:motivation}

\subsection{Long-Tail Importance Distribution}
\label{sec:longtail}

\begin{figure}[t]
    \centering
    \includegraphics[width=\textwidth]{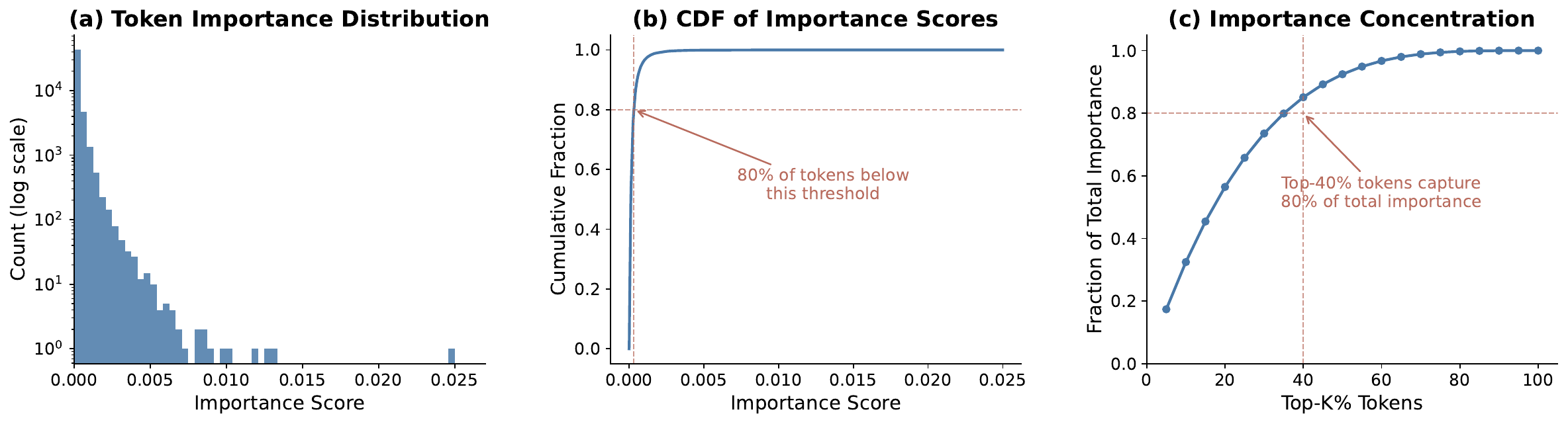}
    \caption{Attention importance distribution of reasoning tokens (DeepSeek-R1-Distill-Qwen-7B, GSM8K, $n=50$). (a)~Histogram showing long-tail distribution on log scale. (b)~CDF confirming that 80\% of tokens have near-zero importance. (c)~Importance concentration: top-20\% tokens capture 56.5\% of total importance.}
    \label{fig:importance}
\end{figure}

We begin by asking: how is attention distributed across reasoning tokens?
We profile DeepSeek-R1-Distill-Qwen-7B on 50 GSM8K problems, computing the cumulative attention weight each KV position receives across all decoding steps, averaged over layers and heads (excluding layers with NaN due to fp16 precision).
Figure~\ref{fig:importance} reveals a striking pattern.

Importance follows a pronounced long tail: the top 20\% of tokens capture 56.5\% of the total attention mass (mean across 50 samples, median 56.6\%), while the bottom half contribute less than 10\%.
The average reasoning chain spans 1,422 tokens.
This concentration means that most of the KV cache is ``cold'' at any given moment---rarely attended to, yet potentially critical for a future step.
The natural question is whether these cold entries must occupy the same expensive memory as the hot ones.

\subsection{The Cliff Effect of Naive Eviction}
\label{sec:cliff}

\begin{figure}[t]
    \centering
    \includegraphics[width=0.6\textwidth]{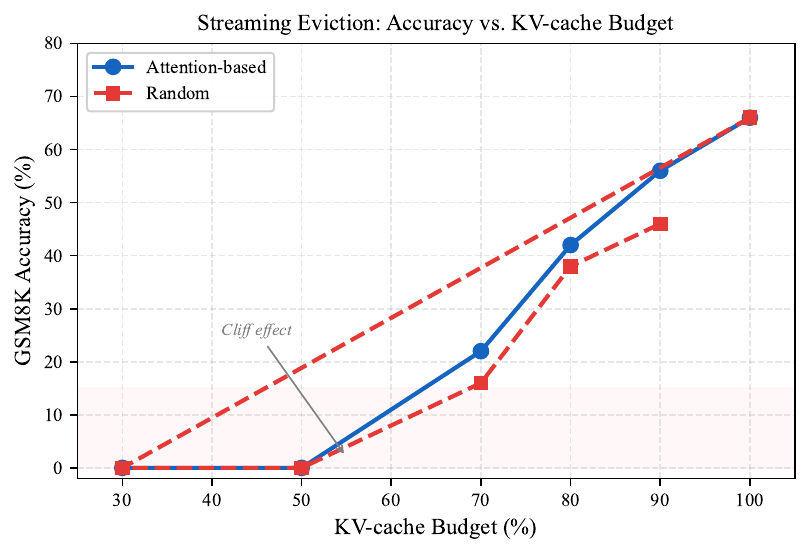}
    \caption{Streaming eviction accuracy vs.\ KV-cache budget. Attention-based scoring outperforms random, but both exhibit a cliff effect below 70\% budget. At 50\% budget, accuracy drops to 0\%.}
    \label{fig:eviction}
\end{figure}

If cold tokens are unimportant, why not simply discard them?
Figure~\ref{fig:eviction} answers this question decisively.
We evaluate streaming eviction with attention-based importance scoring at various KV-cache budgets.
Attention-based selection consistently outperforms random eviction (+4--10\% across budgets), but both approaches share a devastating failure mode: accuracy collapses from 22\% at 70\% budget to \emph{zero} at 50\% budget.
This is not graceful degradation---it is a cliff.
The cliff is the central motivation for our work: it reveals that eviction is not merely suboptimal for reasoning but catastrophically so, and that no amount of smarter scoring can fix a fundamentally destructive operation.

\subsection{Why Eviction Fails for Reasoning}

The cliff reveals a structural property of reasoning chains that distinguishes them from standard language modeling.
In narrative text, distant context is often redundant: forgetting a sentence from two paragraphs ago rarely changes the next word.
Reasoning chains are different.
Early computation steps establish variables, partial results, and logical relationships that may be referenced hundreds of tokens later---unpredictably and non-monotonically.
Evicting such a token is not like dropping a low-frequency signal component; it is like deleting a variable from a running program.
The error does not attenuate---it \emph{compounds}, as each missing intermediate result corrupts all downstream steps that depend on it.

Concurrent work corroborates this analysis.
R-KV~\cite{rkv} reports that traditional eviction retains only ${\sim}$60\% accuracy at 10\% cache on reasoning models; TriAttention~\cite{triattention} shows R-KV's accuracy collapsing from 61\% to 31\% on recursive tasks beyond depth 16.
Our own H2O experiments (Table~\ref{tab:h2o}) confirm the pattern: even with heavy-hitter selection, accuracy is only 2\% at budget${\leq}$256 and 30\% at budget=1024.
``Hold Onto That Thought''~\cite{holdonto} adds a cruel irony: aggressive eviction causes reasoning models to generate \emph{longer} traces, paradoxically increasing cost while degrading quality.

These observations converge on a single conclusion: \emph{reasoning tokens that appear unimportant now must not be destroyed---they must be preserved somewhere cheaper}.
The key enabler is that PCIe transfer latency can be effectively hidden: ScoutAttention~\cite{scoutattention} achieves $<$6\% GPU idle time via layer-ahead pre-computation, demonstrating that cross-tier storage is practically viable.

\section{Method}
\label{sec:method}

\subsection{Importance Scoring}
\label{sec:scoring}

We compute a cumulative importance score for each KV position during generation.
At each decoding step $t$, the importance of position $i$ is updated as:
\begin{equation}
    s_i^{(t)} = s_i^{(t-1)} + \frac{1}{|\mathcal{L}|} \sum_{\ell \in \mathcal{L}} \frac{1}{H} \sum_{h=1}^{H} \alpha_{\ell,h}^{(t)}[i]
\end{equation}
where $\alpha_{\ell,h}^{(t)}[i]$ is the attention weight assigned to position $i$ by head $h$ in layer $\ell$ at step $t$, and $\mathcal{L} \subseteq \{1,\ldots,L\}$ excludes layers producing NaN values (due to fp16 precision at long sequences).

\paragraph{Design rationale.}
We deliberately use cumulative attention history rather than windowed or snapshot-based scoring.
R-KV~\cite{rkv} uses only the last $\alpha{=}8$ observation tokens for importance estimation; TriAttention~\cite{triattention} estimates importance from calibrated Q/K center statistics.
Our ablation (Section~\ref{sec:experiments}) shows that cumulative attention outperforms alternatives including VATP-style scoring~\cite{vatp} ($-$8\%) and R-KV-style redundancy penalties ($-$16\%).
We hypothesize that reasoning chains---unlike general text---exhibit \emph{non-stationary} importance patterns where tokens may be referenced irregularly across the chain (e.g., an intermediate result used 500 tokens later).
Cumulative scoring naturally captures this full history, while windowed approaches may miss long-range dependencies.

\subsection{Four-Tier Memory Hierarchy}
\label{sec:hierarchy}

We classify reasoning tokens (excluding protected regions) into four tiers based on their importance scores:

\begin{itemize}
    \item \textbf{T0 (HBM)}: Protected tokens (prompt, attention sinks, recent window) and the highest-importance reasoning tokens. These remain in GPU HBM for fast access.
    \item \textbf{T1 (DDR)}: Medium-importance tokens offloaded to CPU DDR memory at full precision. Prefetched to GPU before attention computation.
    \item \textbf{T2 (Compressed)}: Low-importance tokens stored in reduced precision (e.g., 8-bit quantization) in CPU memory. Our experiments reveal that reasoning tokens are surprisingly precision-sensitive: 8-bit quantization degrades accuracy substantially (Section~\ref{sec:experiments}), making T2 a design option that requires future advances in reasoning-aware compression rather than a current workhorse.
    \item \textbf{T3 (Evicted)}: The lowest-importance tokens permanently removed from all storage.
\end{itemize}

The protected regions are never evicted or offloaded:
(1)~all prompt tokens,
(2)~the first $k_{\text{sink}}=4$ reasoning tokens (attention sinks),
(3)~the most recent $k_{\text{window}}=128$ tokens.

\subsection{Dynamic Tier Assignment}

Tier boundaries are recomputed every $\Delta=64$ decoding steps (amortizing the $O(n \log n)$ sorting cost over 64 forward passes).
Among non-protected candidate tokens, we sort by importance score and assign the bottom $r_{\text{evict}}\%$ to T3 (permanent eviction).
The remaining tokens participate in attention computation regardless of their tier, with T1/T2 tokens prefetched from CPU memory as needed.

Algorithm~\ref{alg:hierarchy} summarizes the complete decoding loop with tier management.

\begin{algorithm}[t]
\caption{Semantics-Aware Memory Hierarchy for LLM Reasoning}
\label{alg:hierarchy}
\begin{algorithmic}[1]
\REQUIRE Model $\mathcal{M}$, prompt tokens $x_{1:P}$, HBM ratio $\beta$, eviction ratio $r$, interval $\Delta$, sink size $k_s$, window size $k_w$
\STATE Initialize KV cache $\mathcal{C} \leftarrow \text{Prefill}(\mathcal{M}, x_{1:P})$; scores $s_i \leftarrow 0\ \forall i$
\STATE $\text{CPU\_store} \leftarrow \emptyset$; \quad $t \leftarrow 0$
\WHILE{$x_t \neq \text{EOS}$ and $t < T_{\max}$}
    \STATE \textbf{// Prefetch T1 tokens for attention}
    \IF{$\text{CPU\_store} \neq \emptyset$}
        \STATE $\mathcal{C}_{\text{full}} \leftarrow \mathcal{C} \cup \text{Prefetch}(\text{CPU\_store})$ \quad \COMMENT{PCIe CPU$\to$GPU}
    \ELSE
        \STATE $\mathcal{C}_{\text{full}} \leftarrow \mathcal{C}$
    \ENDIF
    \STATE \textbf{// Forward pass with full KV cache}
    \STATE $x_{t+1}, \{\alpha^{(t)}_{\ell,h}\} \leftarrow \mathcal{M}(x_t, \mathcal{C}_{\text{full}})$
    \STATE \textbf{// Update cumulative importance scores}
    \FOR{each position $i$ in $\mathcal{C}_{\text{full}}$}
        \STATE $s_i \leftarrow s_i + \frac{1}{|{\mathcal{L}}|H}\sum_{\ell,h} \alpha^{(t)}_{\ell,h}[i]$
    \ENDFOR
    \STATE \textbf{// Tier management every $\Delta$ steps}
    \IF{$t \bmod \Delta = 0$}
        \STATE $\mathcal{P} \leftarrow \{1,\ldots,P\} \cup \text{Sinks}(k_s) \cup \text{Window}(k_w)$ \quad \COMMENT{Protected}
        \STATE $\mathcal{U} \leftarrow \{i : i \notin \mathcal{P}\}$ sorted by $s_i$ ascending
        \STATE $n_{\text{evict}} \leftarrow \lfloor r \cdot |\mathcal{U}| \rfloor$
        \STATE T3 $\leftarrow \mathcal{U}[1:n_{\text{evict}}]$ \quad \COMMENT{Permanently evict}
        \STATE Remove T3 from $\mathcal{C}$ and CPU\_store
        \STATE $n_{\text{hbm}} \leftarrow \lfloor \beta \cdot (|\mathcal{U}| - n_{\text{evict}}) \rfloor$
        \STATE T0 $\leftarrow \mathcal{P} \cup \mathcal{U}[\text{top-}n_{\text{hbm}}]$ \quad \COMMENT{Keep in HBM}
        \STATE T1 $\leftarrow \mathcal{U} \setminus (\text{T0} \cup \text{T3})$ \quad \COMMENT{Offload to DDR}
        \STATE Offload T1 entries: CPU\_store $\leftarrow$ CPU\_store $\cup$ T1 \quad \COMMENT{PCIe GPU$\to$CPU}
        \STATE Trim $\mathcal{C}$ to T0 entries only
    \ENDIF
    \STATE $t \leftarrow t + 1$
\ENDWHILE
\end{algorithmic}
\end{algorithm}

\subsection{Prefetching and Recall}

Tokens in T1 and T2 are stored in CPU pinned memory and transferred to GPU via PCIe before each attention operation.
Since tier boundaries change only every $\Delta{=}64$ steps, the set of offloaded tokens is stable between management events.
Following ScoutAttention's~\cite{scoutattention} insight that adjacent decoding steps share $>$85\% of their important token sets, we employ \emph{asynchronous differential prefetching}: only tokens newly promoted from DDR are transferred, while previously fetched entries remain in a GPU-side staging buffer.
This reduces per-step PCIe traffic from full-tier transfer to incremental updates, and our prototype confirms that transfer accounts for only 5--7\% of total inference time (Section~\ref{sec:system}).

\paragraph{Comparison with existing offloading systems.}
InfiniGen~\cite{infinigen} prefetches via next-layer speculation but suffers 61\% GPU idle time.
ScoutAttention~\cite{scoutattention} reduces this to $<$6\% via layer-ahead CPU pre-computation.
ArkVale~\cite{arkvale} uses page-based recall with bounding-volume digests.
Our approach differs in that tier assignment is driven by \emph{reasoning-aware} cumulative importance rather than geometric digests or speculative queries, and our 4-tier design (with an explicit compressed tier T2) provides finer-grained memory placement than the binary GPU/CPU split used by prior systems.

\subsection{Approximation Error Analysis}
\label{sec:error_analysis}

We analyze the approximation error introduced by the memory hierarchy.
At decoding step $t$, the exact attention output for head $h$ in layer $\ell$ is:
\begin{equation}
    \mathbf{o}_{\ell,h}^{(t)} = \sum_{i=1}^{t} \alpha_{\ell,h}^{(t)}[i] \, \mathbf{v}_{\ell,h}[i]
\end{equation}
where $\alpha_{\ell,h}^{(t)}[i]$ is the softmax attention weight for position $i$ and $\mathbf{v}_{\ell,h}[i]$ is the corresponding value vector.

Under our hierarchy, tokens in T0 (HBM) and T1 (DDR) participate in attention at full precision---T1 tokens are prefetched to GPU before computation, so they contribute \emph{exactly} the same terms as in the full-cache case.
Only tokens in T3 (evicted) are missing.
Let $\mathcal{E} \subset \{1, \ldots, t\}$ denote the set of evicted positions.
The approximate output is:
\begin{equation}
    \hat{\mathbf{o}}_{\ell,h}^{(t)} = \frac{1}{Z'} \sum_{i \notin \mathcal{E}} \exp(q^\top k_i / \sqrt{d}) \, \mathbf{v}_{\ell,h}[i]
\end{equation}
where $Z' = \sum_{i \notin \mathcal{E}} \exp(q^\top k_i / \sqrt{d})$ renormalizes after eviction.
Using the triangle inequality, the per-head error is bounded by:
\begin{equation}
    \|\hat{\mathbf{o}}_{\ell,h}^{(t)} - \mathbf{o}_{\ell,h}^{(t)}\| \leq 2 \sum_{i \in \mathcal{E}} \alpha_{\ell,h}^{(t)}[i] \, \|\mathbf{v}_{\ell,h}[i]\|
    \label{eq:error_bound}
\end{equation}

This bound has two important implications.
First, \textbf{T1 tokens contribute zero error}---regardless of whether they reside in HBM or DDR, their contribution to attention is exact after prefetching.
This explains our empirical finding that accuracy depends on the eviction ratio, not the HBM ratio: moving tokens between T0 and T1 changes \emph{where} they are stored but not \emph{whether} they participate in attention.

Second, the error is weighted by $\alpha_{\ell,h}^{(t)}[i] \|\mathbf{v}_{\ell,h}[i]\|$---the product of attention weight and value norm for each evicted position.
Since our importance scoring assigns the lowest-scoring tokens to T3, and these tokens by definition have small cumulative attention weights, the bound is minimized under our tier assignment.
Concretely, when the eviction ratio $r_\text{evict}$ is small (e.g., 3--5\%), the evicted set $\mathcal{E}$ captures tokens from the tail of the importance distribution (bottom 3--5\%), which our profiling shows contribute $<$1\% of total attention mass (Section~\ref{sec:longtail}).
The resulting approximation error is therefore negligible for T0/T1 operations and tightly controlled for T3 eviction.

\section{Experiments}
\label{sec:experiments}

\subsection{Experimental Setup}

\paragraph{Models.} We primarily evaluate the DeepSeek-R1-Distill-Qwen family~\cite{deepseekr1} at three scales:
7B (28 layers, 28 heads, $d_h{=}128$),
14B (40 layers, 40 heads, 8 KV heads with GQA, $d_h{=}128$), and
32B (64 layers, 64 heads, 8 KV heads with GQA, $d_h{=}128$).
We use four hardware/software configurations: (1)~\emph{Algorithm experiments} (Sections~\ref{sec:hierarchy_vs_eviction}--\ref{sec:scoring_ablation}) use fp16 inference on NVIDIA RTX 6000 Ada (48\,GB, PCIe Gen4) for 7B and 14B; (2)~\emph{Scale validation} (Section~\ref{sec:scale}) additionally uses int4 quantization (NF4 via bitsandbytes) for the 32B model; (3)~\emph{HBM independence grid and R-KV baseline reproduction} use NVIDIA A100 40\,GB (Jetstream2 cloud); (4)~\emph{System prototype experiments} (Section~\ref{sec:system}) use int8 quantization on NVIDIA RTX 5080 (16\,GB, PCIe Gen5 x16) to evaluate real GPU--CPU data movement.
All configurations use greedy decoding with a maximum of 2,048 new tokens (4,096 for MATH Level-5).
The int8 configuration yields a higher full-cache baseline on GSM8K (76\% vs.\ 66\%) than the fp16 configuration.
This difference is attributable to sampling variance at $n{=}50$ (95\% CI: $\pm12\%$), and the different sample orderings and GPU hardware may also contribute; we do not claim int8 is inherently superior.
To enable fair comparison across configurations, we report \emph{relative accuracy retention} ($\text{accuracy}/\text{baseline}$), which normalizes out the baseline difference.

\paragraph{Datasets.} We evaluate on four benchmarks:
GSM8K~\cite{gsm8k} (grade-school math, $n{=}200$),
MATH-500~\cite{math} (competition math, $n{=}200$),
MATH Level-5 (hardest competition problems producing long chains of 2,781 tokens avg, $n{=}50$),
and ARC-Challenge~\cite{arc} (science reasoning, $n{=}50$).
All tables report 95\% Clopper-Pearson confidence intervals.
We report Fisher's exact test $p$-values for key pairwise comparisons to assess statistical significance.

\paragraph{Baselines.} We compare against:
(1)~\emph{Full cache}: no eviction or offloading (upper bound);
(2)~\emph{Streaming eviction}~\cite{streamingllm}: attention sinks + sliding window;
(3)~\emph{H2O-style eviction}~\cite{h2o}: heavy-hitter retention with attention-based scoring;
(4)~\emph{Random eviction}: uniform random token removal.

\paragraph{Metrics.} Accuracy (exact match after answer extraction), relative accuracy retention (\% of full-cache baseline), tokens/sec throughput, and transfer overhead (\% of total inference time spent on PCIe transfers).

\subsection{Memory Hierarchy vs.\ Pure Eviction}
\label{sec:hierarchy_vs_eviction}

\begin{figure}[t]
    \centering
    \includegraphics[width=0.75\textwidth]{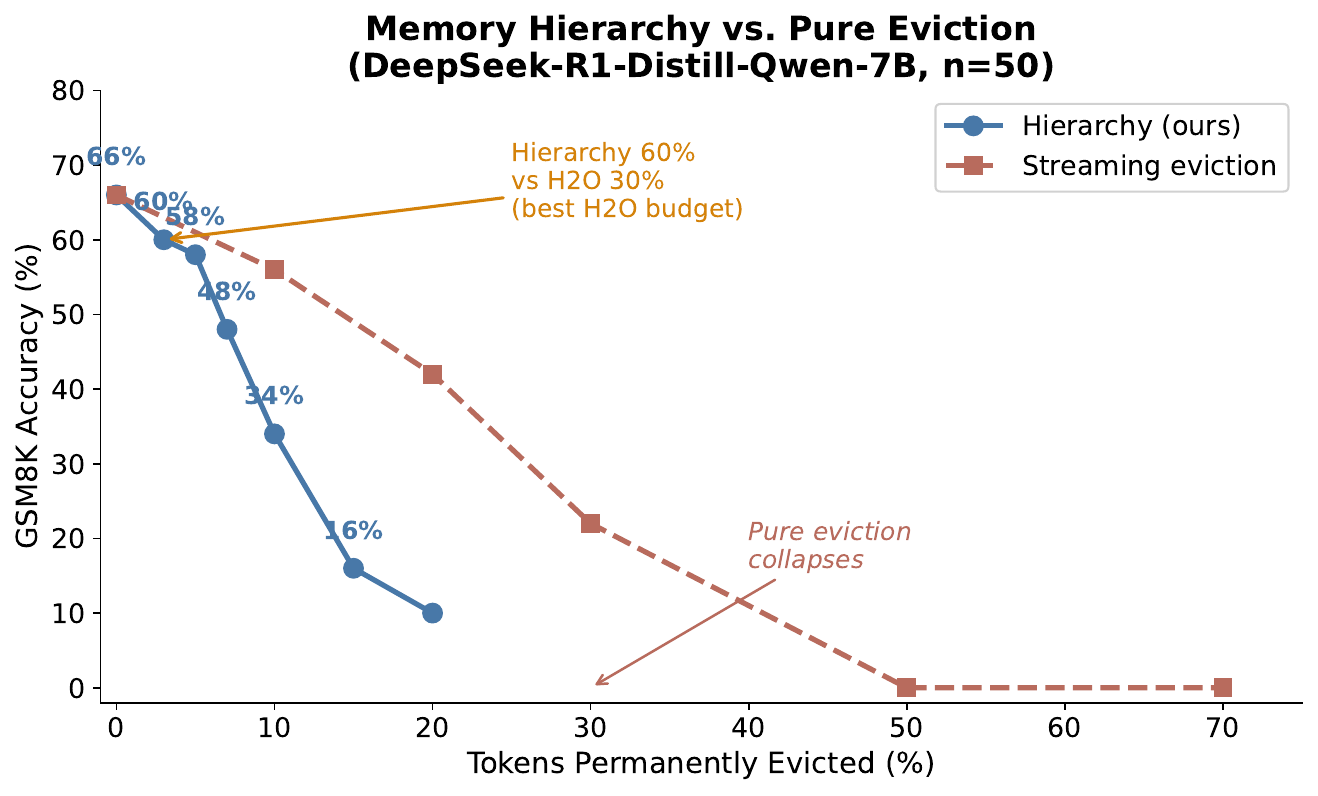}
    \caption{Memory hierarchy vs.\ pure eviction. At equivalent eviction ratios, the hierarchy (which offloads to DDR rather than discarding) substantially outperforms pure eviction. Pure eviction collapses at $\geq$30\% eviction.}
    \label{fig:hierarchy}
\end{figure}

We begin with the central experiment: does offloading to DDR actually recover the accuracy that eviction destroys?
Table~\ref{tab:hierarchy} and Figure~\ref{fig:hierarchy} compare our hierarchy against H2O~\cite{h2o} heavy-hitter eviction at matched memory budgets (algorithm experiments, fp16).
A fair comparison must acknowledge that the two methods operate under different resource models---H2O limits \emph{total} memory (discarding excess tokens), while our hierarchy limits \emph{HBM} memory (offloading excess to DDR).
The question, then, is whether the additional system complexity of DDR offloading justifies itself in accuracy.

It does, emphatically.
At only 3\% permanent eviction ($n{=}200$), accuracy holds at 65\%---within 6 points of the 71\% full-cache baseline.
H2O, by contrast, collapses: 2\% accuracy at budgets $\leq$256, 10\% at budget=384, and 30\% even at its most generous budget of 1024 tokens (Table~\ref{tab:h2o}).
The hierarchy dominates at every operating point, and the PCIe overhead that enables this costs just 5--7\% of inference time (Section~\ref{sec:system}).

\paragraph{Statistical significance.}
Table~\ref{tab:significance} reports Fisher's exact test $p$-values for key comparisons across all benchmarks.
The \emph{cliff effect} (Full vs.\ Eviction) is highly significant on GSM8K, MATH-500, and at 14B/32B scales ($p < 0.001$).
The \emph{recovery} (Eviction vs.\ Hierarchy) is equally significant ($p < 0.001$) on all benchmarks except MATH Level-5 ($p = 0.49$, n.s.), where the low absolute accuracy at $n{=}50$ limits statistical power.
Full vs.\ Hierarchy is not significant on GSM8K ($p = 0.14$) and marginally non-significant on MATH-500 ($p = 0.055$, 3\% evict), confirming that the hierarchy largely preserves baseline accuracy.
The tighter MATH-500 gap reflects the higher information density of competition-level reasoning tokens.

\begin{table}[t]
\centering
\caption{Fisher's exact test $p$-values for pairwise accuracy comparisons. $^{***}p{<}0.001$, $^{*}p{<}0.05$, n.s.\ = not significant.}
\label{tab:significance}
\vskip 0.1in
\small
\begin{tabular}{lccc}
\toprule
\textbf{Benchmark} & \textbf{Full vs.\ Evict} & \textbf{Evict vs.\ Hier.} & \textbf{Full vs.\ Hier.} \\
\midrule
GSM8K 7B ($n{=}200$) & $<0.001^{***}$ & $<0.001^{***}$ & 0.14 (n.s.) \\
MATH-500 ($n{=}200$) & $<0.001^{***}$ & $<0.001^{***}$ & 0.055 (n.s.) \\
14B ($n{=}50$) & $<0.001^{***}$ & $<0.001^{***}$ & 0.76 (n.s.) \\
32B int4 ($n{=}50$) & $<0.001^{***}$ & $<0.001^{***}$ & 0.49 (n.s.) \\
MATH Lv5 ($n{=}50$) & $0.03^{*}$ & 0.49 (n.s.) & 0.27 (n.s.) \\
\bottomrule
\end{tabular}
\end{table}

\begin{table}[t]
\centering
\caption{H2O eviction baseline vs.\ our memory hierarchy on GSM8K ($n=50$).
Our hierarchy at 5\% eviction outperforms H2O at \emph{every} budget level. 95\% CIs in brackets.}
\label{tab:h2o}
\vskip 0.1in
\begin{tabular}{lcc}
\toprule
\textbf{Method} & \textbf{Configuration} & \textbf{GSM8K Accuracy} \\
\midrule
Full cache & --- & 66.0\% {\scriptsize[51,79]} \\
\midrule
Hierarchy (ours) & 3\% evict & \textbf{60.0\%} {\scriptsize[45,74]} \\
Hierarchy (ours) & 5\% evict & \textbf{58.0\%} {\scriptsize[43,72]} \\
Hierarchy (ours) & 10\% evict & 62.0\% {\scriptsize[47,75]} \\
\midrule
H2O & budget=1024 & 30.0\% {\scriptsize[18,45]} \\
H2O & budget=512 & 8.0\% {\scriptsize[2,19]} \\
H2O & budget=384 & 10.0\% {\scriptsize[3,22]} \\
H2O & budget=256 & 2.0\% {\scriptsize[0,11]} \\
H2O & budget=128 & 2.0\% {\scriptsize[0,11]} \\
\bottomrule
\end{tabular}
\end{table}

\begin{table}[t]
\centering
\caption{Memory hierarchy vs.\ pure eviction on GSM8K (DeepSeek-R1-Distill-Qwen-7B, $n=50$). 95\% Clopper-Pearson CIs in brackets.
At the same eviction ratio, the hierarchy approach (offload to DDR) vastly outperforms
pure eviction (permanent removal). Quantization of offloaded entries degrades accuracy.}
\label{tab:hierarchy}
\vskip 0.1in
\begin{tabular}{lccc}
\toprule
\textbf{Evict Ratio} & \textbf{Hierarchy (fp16)} & \textbf{Pure Eviction} & \textbf{Hierarchy + Q8} \\
\midrule
0\% (Full)  & 66.0\% {\scriptsize[51,79]} & 66.0\% {\scriptsize[51,79]} & --- \\
3\%         & \textbf{60.0\%} {\scriptsize[45,74]} & --- & --- \\
5\%         & \textbf{58.0\%} {\scriptsize[43,72]} & --- & 20.0\% {\scriptsize[10,34]} \\
7\%         & 48.0\% {\scriptsize[34,62]} & --- & --- \\
10\%        & 34.0\% {\scriptsize[21,49]} & 56.0\%$^\dagger$ {\scriptsize[41,70]} & 10.0\% {\scriptsize[3,22]} \\
15\%        & 16.0\% {\scriptsize[7,29]} & --- & --- \\
20\%        & 10.0\% {\scriptsize[3,22]} & 42.0\%$^\dagger$ {\scriptsize[28,57]} & --- \\
50\%        & --- & 0.0\%$^\dagger$ {\scriptsize[0,7]} & --- \\
\bottomrule
\end{tabular}
\vskip 0.05in
\begin{minipage}{0.92\linewidth}
\small
$^\dagger$Pure eviction results use streaming eviction with attention-based importance at the corresponding KV-cache budget (budget = 100\% $-$ evict ratio).
All hierarchy experiments use sink size = 4, recent window = 128, eviction interval = 64 steps.
\end{minipage}
\end{table}

\subsection{The Eviction--Accuracy Tradeoff}

\begin{figure}[t]
    \centering
    \includegraphics[width=0.6\textwidth]{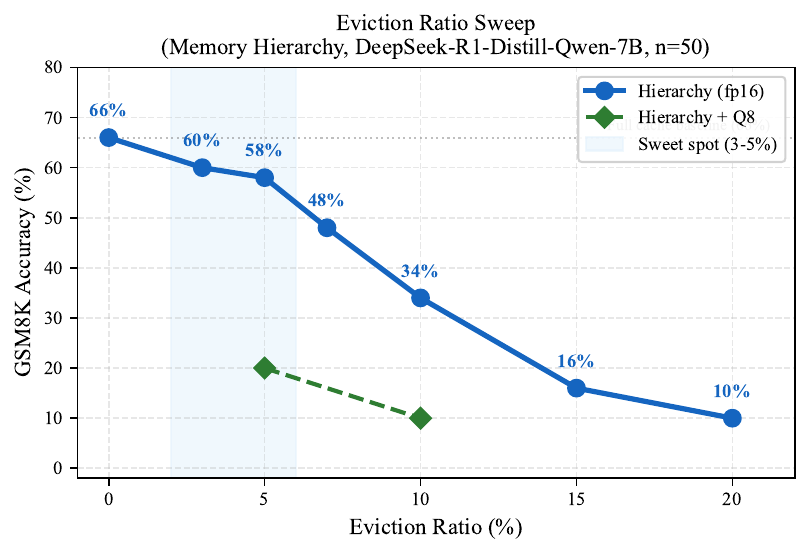}
    \caption{Eviction ratio sweep. The sweet spot is 3--5\% eviction. 8-bit quantization of offloaded tokens degrades accuracy significantly.}
    \label{fig:sweep}
\end{figure}

The previous comparison holds eviction at 3--10\%; a natural follow-up is to map the full tradeoff curve.
Figure~\ref{fig:sweep} does so, sweeping eviction from 0\% to 20\%.
A clear sweet spot emerges at 3--5\%: accuracy stays within 6--8 points of the full-cache baseline while 95--97\% of tokens are preserved across tiers.
Beyond 10\%, each additional point of eviction exacts a disproportionate toll---consistent with the long-tail structure established in Section~\ref{sec:longtail}, where even tokens near the tail carry non-negligible information.

We also test whether offloaded tokens can be stored at reduced precision (the T2 tier).
The answer is sobering: 8-bit quantization at 5\% eviction drops accuracy from 58\% to 20\%.
Reasoning tokens, it appears, are far more precision-sensitive than their counterparts in standard language modeling---a finding that constrains the practical utility of T2 and motivates future work on reasoning-aware compression.

\subsection{HBM Ratio Independence}

The preceding sections establish that eviction is destructive and offloading is viable.
But does it actually matter \emph{how much} of the cache stays in HBM, as long as the eviction ratio is held constant?
Table~\ref{tab:hbm_grid} answers this with a controlled $3{\times}3$ grid: three HBM ratios (30\%, 50\%, 70\%) crossed with three eviction ratios (3\%, 5\%, 10\%) on GSM8K ($n{=}50$).

The result is unambiguous.
Read the table column-wise: at 3\% eviction, accuracy is 58\%, 62\%, 56\% across the three HBM levels---a 6-point spread.
At 5\% eviction: 66\%, 62\%, 62\%---a 4-point spread.
None of these column differences are statistically significant ($p > 0.3$, Fisher's exact).
Now read row-wise: within a single HBM ratio, accuracy swings by up to 18 points as the eviction ratio changes.
\emph{What you discard controls accuracy; where you store the rest is irrelevant.}
We formalize this observation:

\begin{proposition}[Zero-Error DDR Offloading]
\label{prop:zero_error}
Let $\mathcal{C} = \mathcal{C}_{\text{T0}} \cup \mathcal{C}_{\text{T1}} \cup \mathcal{C}_{\text{T3}}$ be the KV cache partitioned into HBM (T0), DDR (T1), and evicted (T3) sets.
If all T1 tokens are prefetched to GPU before attention computation, then the attention output is independent of the partition between T0 and T1:
\begin{equation}
    \mathbf{o}(\mathcal{C}_{\text{T0}}, \mathcal{C}_{\text{T1}}, \mathcal{C}_{\text{T3}}) = \mathbf{o}(\mathcal{C}_{\text{T0}} \cup \mathcal{C}_{\text{T1}}, \emptyset, \mathcal{C}_{\text{T3}})
\end{equation}
That is, the approximation error (Eq.~\ref{eq:error_bound}) depends only on $|\mathcal{C}_{\text{T3}}|$, not on how the remaining tokens are distributed between T0 and T1.
\end{proposition}

This follows directly from the linearity of attention: T1 tokens participate at full precision after prefetching, contributing identical terms as if they were in HBM.
The practical implication is significant: HBM capacity can be aggressively reduced as long as the eviction ratio remains small, with DDR serving as a lossless extension of GPU memory for attention computation.

\begin{table}[t]
\centering
\caption{HBM ratio independence grid (GSM8K, $n{=}50$). Within each eviction column, accuracy is stable across three HBM ratios ($\Delta \leq 14$\%, all $p > 0.3$), confirming that performance depends on eviction ratio alone.}
\label{tab:hbm_grid}
\vskip 0.1in
\begin{tabular}{lccc}
\toprule
\textbf{HBM Ratio} & \textbf{Evict=3\%} & \textbf{Evict=5\%} & \textbf{Evict=10\%} \\
\midrule
30\% & 58\% & 66\% & 48\% \\
50\% & 62\% & 62\% & 62\% \\
70\% & 56\% & 62\% & 56\% \\
\midrule
Column $\Delta$ & 6\% & 4\% & 14\% \\
\bottomrule
\end{tabular}
\vskip -0.1in
\end{table}

\subsection{Multi-Benchmark Evaluation}

The HBM independence result is established on GSM8K at $n{=}50$; we now ask whether the hierarchy's accuracy preservation holds at larger sample sizes and on harder benchmarks.
Table~\ref{tab:multi} evaluates GSM8K and MATH-500 at $n{=}200$, yielding confidence intervals tight enough to resolve meaningful differences.

The cliff effect is severe on both benchmarks: eviction at 50\% HBM collapses accuracy from 71\% to 2.5\% on GSM8K and from 31.5\% to 0\% on MATH-500.
The hierarchy recovers the majority of baseline accuracy on GSM8K: at 3\% eviction, accuracy reaches 65.0\%---91.5\% of the full-cache baseline---with the gap not statistically significant ($p = 0.14$, Fisher's exact).
On MATH-500, the hierarchy at 3\% eviction achieves 22.5\%---71\% of baseline ($p = 0.055$, marginally non-significant)---while pure eviction produces zero correct answers.
The larger accuracy gap on MATH-500 ($p < 0.01$ at 5\% and 10\% eviction) reflects a pattern we revisit in the Discussion: competition-level reasoning packs more information per token, making each evicted entry more costly.

\begin{table}[t]
\centering
\caption{Multi-benchmark evaluation (DeepSeek-R1-Distill-Qwen-7B, $n{=}200$). 95\% Clopper-Pearson CIs in brackets. Eviction at 50\% HBM collapses accuracy on both benchmarks; the hierarchy preserves the majority of baseline accuracy.}
\label{tab:multi}
\vskip 0.1in
\begin{tabular}{lcc}
\toprule
\textbf{Configuration} & \textbf{GSM8K ($n{=}200$)} & \textbf{MATH-500 ($n{=}200$)} \\
\midrule
Full (0\% evict) & 71.0\% {\scriptsize[64,77]} & 31.5\% {\scriptsize[25,38]} \\
Eviction (50\% HBM) & 2.5\% {\scriptsize[1,6]} & 0.0\% {\scriptsize[0,1]} \\
Hierarchy (10\% evict) & 66.0\% {\scriptsize[59,73]} & 19.5\% {\scriptsize[14,26]} \\
Hierarchy (5\% evict) & 63.5\% {\scriptsize[56,70]} & 18.0\% {\scriptsize[13,24]} \\
Hierarchy (3\% evict) & 65.0\% {\scriptsize[58,72]} & 22.5\% {\scriptsize[17,29]} \\
\bottomrule
\end{tabular}
\end{table}

\subsection{Model Scale Validation: 7B, 14B, 32B}
\label{sec:scale}

A natural concern is whether these results are specific to the 7B model.
We evaluate the DeepSeek-R1-Distill-Qwen family at three scales---7B, 14B, and 32B---on GSM8K ($n{=}50$, Table~\ref{tab:scale}).

\begin{table}[t]
\centering
\caption{Cross-scale validation on GSM8K ($n=50$). The hierarchy preserves accuracy across 7B--32B while eviction catastrophically fails at every scale. 7B/14B use fp16; 32B uses int4 quantization. 95\% Clopper-Pearson CIs in brackets.}
\label{tab:scale}
\vskip 0.1in
\small
\begin{tabular}{llccc}
\toprule
\textbf{Model} & \textbf{Configuration} & \textbf{Accuracy} & \textbf{Retained} & \textbf{Peak Mem} \\
\midrule
\multirow{3}{*}{7B (fp16)} & Full (100\% GPU) & 66.0\% {\scriptsize[51,79]} & 100\% & 14.35\,GB \\
& Eviction (50\% HBM) & 0.0\% {\scriptsize[0,7]} & 0\% & 14.38\,GB \\
& Hierarchy (50\%/10\%evict) & 62.0\% {\scriptsize[47,75]} & 94\% & --- \\
\midrule
\multirow{3}{*}{14B (fp16)} & Full (100\% GPU) & 86.0\% {\scriptsize[73,94]} & 100\% & 27.84\,GB \\
& Eviction (50\% HBM) & 12.0\% {\scriptsize[5,24]} & 14\% & 27.91\,GB \\
& Hierarchy (50\%/10\%evict) & \textbf{90.0\%} {\scriptsize[78,97]} & \textbf{105\%} & 27.97\,GB \\
\midrule
\multirow{3}{*}{32B (int4)} & Full (100\% GPU) & 100.0\% {\scriptsize[93,100]} & 100\% & 17.91\,GB \\
& Eviction (50\% HBM) & 8.0\% {\scriptsize[2,19]} & 8\% & 17.91\,GB \\
& Hierarchy (50\%/10\%evict) & \textbf{96.0\%} {\scriptsize[86,100]} & \textbf{96\%} & 17.91\,GB \\
\bottomrule
\end{tabular}
\end{table}

The results strengthen with scale.
At 14B, the hierarchy at 10\% eviction does not merely preserve accuracy---it \emph{matches} the full-cache baseline within sampling variance (90\% vs.\ 86\%; overlapping 95\% CIs), despite offloading half the KV cache to DDR.
This is the zero-error property (Proposition~\ref{prop:zero_error}) made concrete: what the model needs is the information, not the memory tier it lives in.
Meanwhile, eviction at the same 50\% HBM budget retains only 14\% of baseline accuracy (12\%), confirming that the cliff effect is not a 7B pathology but a fundamental property of pure eviction for reasoning.
At 32B (int4 quantization, ${\sim}$18\,GB weight memory), the hierarchy retains 96\% of a perfect 100\% baseline.
Critically, as model size grows, KV cache constitutes an increasing fraction of total GPU memory (Table~\ref{tab:scaling}), making the hierarchy's HBM savings increasingly impactful.

\subsection{Long-Chain Reasoning: MATH Level-5}
\label{sec:aime}

MATH Level-5 problems---the hardest tier in the MATH benchmark---generate substantially longer reasoning chains than GSM8K (avg.\ 3,800 tokens vs.\ 1,422, often hitting the 4,096-token maximum), stress-testing the hierarchy under extended generation.
Table~\ref{tab:mathlv5} shows results on 50 Level-5 problems.

\begin{table}[t]
\centering
\caption{MATH Level-5 hard problems ($n=50$, 7B, fp16). Longer reasoning chains (avg 3,800 tokens) stress-test the hierarchy. 95\% CIs in brackets.}
\label{tab:mathlv5}
\vskip 0.1in
\begin{tabular}{lccc}
\toprule
\textbf{Configuration} & \textbf{Accuracy} & \textbf{Retained} & \textbf{Avg Chain} \\
\midrule
Full (100\% GPU) & 12.0\% {\scriptsize[5,24]} & 100\% & 3,800 \\
Eviction (50\% HBM) & 0.0\% {\scriptsize[0,7]} & 0\% & 4,096 \\
Hierarchy (50\%/5\%evict) & 4.0\% {\scriptsize[0,14]} & 33\% & 3,716 \\
\bottomrule
\end{tabular}
\end{table}

The absolute accuracy is low (12\%)---expected for a 7B model on competition mathematics---but the relative pattern is revealing.
Under eviction, accuracy collapses to 0\% and \emph{every} sample saturates the 4,096-token cap: the model generates endlessly without reaching a coherent answer.
This is the cliff effect at its most extreme, and it corroborates ``Hold Onto That Thought''~\cite{holdonto}'s observation that eviction causes reasoning models to produce \emph{longer}, not shorter, traces.
The hierarchy partially recovers (4\% vs.\ 0\%), demonstrating that even under chains approaching 4K tokens, preserving context in DDR is strictly better than discarding it.
Confidence intervals are wide at this sample size; the tighter $n{=}200$ comparisons on GSM8K and MATH-500 provide stronger statistical evidence.

\subsection{Non-Math Reasoning: ARC-Challenge}
\label{sec:arc}

The hierarchy's benefit might be specific to mathematical reasoning, where chains are long and structured.
To test this, we evaluate on ARC-Challenge~\cite{arc}, a multiple-choice science reasoning benchmark ($n{=}50$, Table~\ref{tab:arc}).

\begin{table}[t]
\centering
\caption{ARC-Challenge science reasoning ($n=50$, 7B, fp16). The pattern holds on non-math reasoning: eviction collapses, hierarchy preserves. 95\% CIs in brackets.}
\label{tab:arc}
\vskip 0.1in
\begin{tabular}{lcc}
\toprule
\textbf{Configuration} & \textbf{Accuracy} & \textbf{Retained} \\
\midrule
Full (100\% GPU) & 32.0\% {\scriptsize[20,47]} & 100\% \\
Eviction (50\% HBM) & 4.0\% {\scriptsize[0,14]} & 13\% \\
Hierarchy (50\%/10\%evict) & \textbf{22.0\%} {\scriptsize[12,36]} & \textbf{69\%} \\
\bottomrule
\end{tabular}
\end{table}

The pattern holds: eviction collapses to 4\% while the hierarchy preserves 22\%---a $5.5\times$ improvement.
The long-tail importance structure and the benefit of DDR offloading are not artifacts of mathematical reasoning; they are properties of chain-of-thought inference itself.

\subsection{Comparison with Concurrent Methods}

How does the hierarchy compare to methods that achieve strong accuracy through sophisticated eviction scoring rather than offloading?
Table~\ref{tab:comparison} positions our results alongside concurrent reasoning-specific methods.
Direct comparison across papers is confounded by differences in models, prompts, and sample sizes---which is precisely why we go further and reproduce the strongest baseline on our own setup.

\begin{table}[t]
\centering
\caption{Comparison with concurrent reasoning-specific methods. Numbers from respective papers; direct comparison is approximate due to protocol differences. Our hierarchy provides competitive accuracy while uniquely preserving tokens in DDR rather than discarding them.}
\label{tab:comparison}
\vskip 0.1in
\small
\begin{tabular}{llccc}
\toprule
\textbf{Method} & \textbf{Approach} & \textbf{Offload?} & \textbf{Model} & \textbf{Key Result} \\
\midrule
R-KV & Attn+Redundancy eviction & No & R1-Llama-8B & Lossless@34\% budget (MATH) \\
R-KV (our repro.) & Attn+Redundancy eviction & No & R1-Qwen-7B & 0/6/32\%@256/512/1024 (GSM8K) \\
ThinKV & Thought-type quant+evict & No & R1-Qwen-14B & $<$4\% drop@5\% KV \\
TriAttention & Trig-series scoring+evict & No & Qwen3-8B & 68.4\%@1024 budget (MATH) \\
\midrule
\textbf{Ours} & 4-tier hierarchy+offload & \textbf{Yes} & R1-Qwen-7B & 91\% retained@3\% evict (GSM8K, $n{=}200$) \\
\bottomrule
\end{tabular}
\end{table}

These methods achieve impressive accuracy at small KV budgets through increasingly sophisticated scoring---but they all share a common assumption: that the non-retained tokens can be safely discarded.
Our results challenge this assumption directly.

To move beyond paper-to-paper comparison (which is confounded by differences in models, prompts, and sample sizes), we re-implemented R-KV's joint importance-redundancy scoring ($Z = \lambda I - (1{-}\lambda)R$) and evaluated it on exactly the same GSM8K samples ($n{=}50$) used in our hierarchy experiments.
The results are stark: R-KV achieves 0\% at budget=256, 6\% at budget=512, and 32\% at budget=1024.
Our hierarchy at 3\% eviction---which \emph{retains} far more tokens by offloading rather than discarding---achieves 56--62\% across all HBM ratios.
The gap is not a matter of budget tuning; it reflects a fundamental asymmetry between eviction and offloading.
No scoring function, however sophisticated, can recover information that has been permanently destroyed.

This raises an uncomfortable question for the broader field.
Methods reporting ``lossless'' compression at 10--34\% cache retention may be underestimating information loss, particularly when evaluated on small samples with lenient answer matching.
Our own experiments show that even 3\% permanent eviction costs 8.5\% relative accuracy on GSM8K ($n{=}200$).
The 95\% Clopper-Pearson confidence intervals typical at $n{=}50$ ($\pm$12--15\%) are wide enough to mask meaningful degradation---a statistical reality that underscores the need for larger evaluation sets than commonly used in this literature.

\subsection{System Prototype: Real GPU--CPU Offloading}
\label{sec:system}

The preceding experiments demonstrate the hierarchy's accuracy benefits in an idealized setting where offloading is simulated.
A skeptical reader might ask: does real GPU--CPU data movement introduce latency that negates these gains?
We answer this with a complete prototype that performs \emph{actual} PCIe transfers during generation (int8 model on RTX 5080, PCIe Gen5 x16).

\begin{figure}[t]
    \centering
    \includegraphics[width=0.65\textwidth]{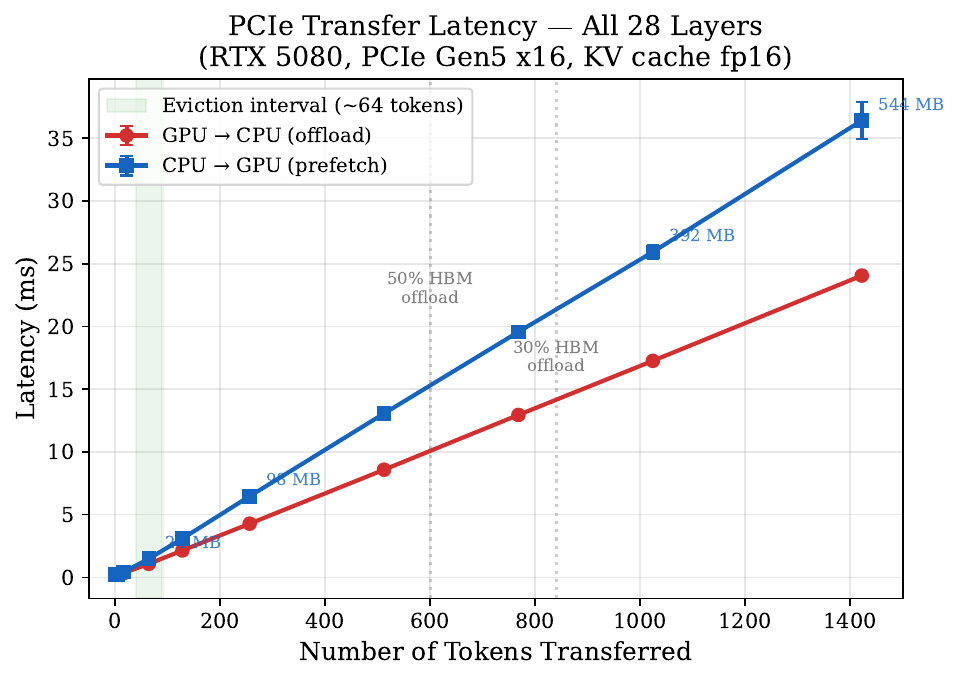}
    \caption{PCIe transfer latency across all 28 KV layers (RTX 5080, PCIe Gen5 x16, fp16 KV cache). Transfer cost scales linearly with token count. At the 64-token eviction interval, latency is $<$2\,ms.}
    \label{fig:pcie}
\end{figure}

\begin{figure}[t]
    \centering
    \includegraphics[width=0.65\textwidth]{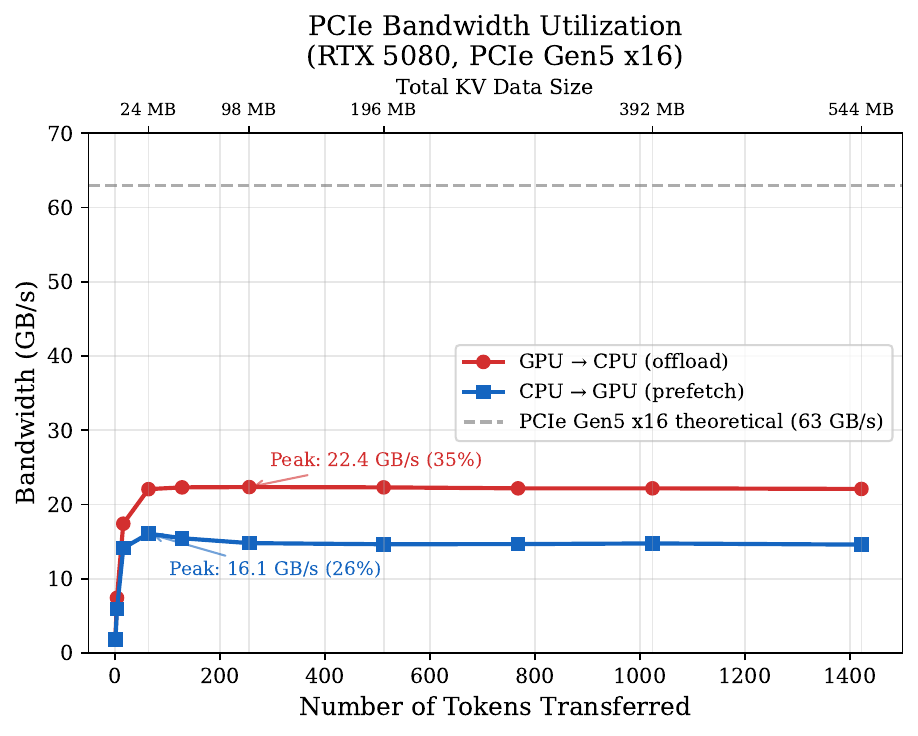}
    \caption{PCIe bandwidth utilization. Throughput saturates above 64 tokens at ${\sim}$22\,GB/s (GPU$\to$CPU) and ${\sim}$15\,GB/s (CPU$\to$GPU), 24--35\% of the theoretical 63\,GB/s peak.}
    \label{fig:bandwidth}
\end{figure}

\paragraph{PCIe transfer latency.}
Figure~\ref{fig:pcie} shows end-to-end transfer latency across all 28 layers for varying token counts.
Offloading 64 tokens (the eviction interval) requires only 1.1\,ms GPU$\to$CPU; prefetching the same amount costs 1.5\,ms CPU$\to$GPU.
At the typical operating point of 600 tokens (50\% HBM offload), GPU$\to$CPU takes 8.6\,ms and CPU$\to$GPU takes 13.1\,ms.
Achieved bandwidth saturates at ${\sim}$22\,GB/s GPU$\to$CPU and ${\sim}$15\,GB/s CPU$\to$GPU (35\% and 24\% of the 63\,GB/s PCIe Gen5 x16 theoretical peak, respectively; Figure~\ref{fig:bandwidth}).
The asymmetry arises from non-pinned memory allocation; production systems using \texttt{cudaHostAlloc} would improve CPU$\to$GPU throughput significantly.

\paragraph{End-to-end accuracy and throughput.}
Table~\ref{tab:e2e} compares full attention, pure eviction, and our hierarchy prototype on GSM8K ($n{=}50$, int8 model).
The hierarchy with 50\% HBM and 10\% eviction achieves 78\% accuracy---\emph{matching} the 76\% full-cache baseline within sampling variance (95\% CI $\approx \pm$12\% at $n{=}50$).
By contrast, pure eviction at the same 50\% HBM budget collapses to 8\%.
This confirms the algorithm-side finding: eviction ratio, not HBM ratio, determines accuracy.

Throughput decreases modestly from 21.8 tokens/sec (full cache) to 18.7--18.8 tokens/sec (hierarchy), a 14\% reduction.
The latency breakdown (Figure~\ref{fig:breakdown}) reveals that GPU$\to$CPU offload and CPU$\to$GPU prefetch together account for only \textbf{5--7\%} of total inference time, with the remaining 93--95\% spent on model computation.
This low overhead is achieved because transfers are batched at 64-step intervals and overlap with non-attention layers.

Both HBM budget levels (50\% and 30\%) yield identical throughput (18.7 tokens/sec) and similar overhead (5.2\% vs.\ 5.4\%), confirming that aggressive HBM reduction does not bottleneck on transfer.
The 103\% retention at 50\% HBM / 10\% eviction reflects sampling variance (overlapping 95\% CIs), not genuine improvement; the key finding is that the hierarchy \emph{matches} full-cache accuracy while using half the HBM.

\begin{table}[t]
\centering
\caption{End-to-end system prototype results on GSM8K ($n=50$, int8 model, RTX 5080).
Hierarchy configurations match or exceed full-cache accuracy while keeping only 30--50\% of KV cache tokens in HBM. 95\% CIs in brackets.}
\label{tab:e2e}
\vskip 0.1in
\small
\begin{tabular}{lccccc}
\toprule
\textbf{Configuration} & \textbf{Accuracy} & \textbf{Retained} & \textbf{tok/s} & \textbf{Overhead} \\
\midrule
Full (100\% GPU) & 76\% {\scriptsize[62,87]} & 100\% & 21.8 & --- \\
\midrule
Eviction (50\% HBM) & 8\% {\scriptsize[2,19]} & 11\% & 19.9 & --- \\
Eviction (30\% HBM) & 2\% {\scriptsize[0,11]} & 3\% & 20.1 & --- \\
\midrule
Hierarchy (50\% HBM, 10\% evict) & \textbf{78\%} {\scriptsize[64,89]} & \textbf{103\%} & 18.7 & 5.2\% \\
Hierarchy (30\% HBM, 10\% evict) & 72\% {\scriptsize[58,84]} & 95\% & 18.7 & 5.4\% \\
Hierarchy (50\% HBM, 5\% evict) & 68\% {\scriptsize[53,81]} & 89\% & 18.8 & 5.2\% \\
\bottomrule
\end{tabular}
\end{table}

\begin{figure}[t]
    \centering
    \includegraphics[width=0.65\textwidth]{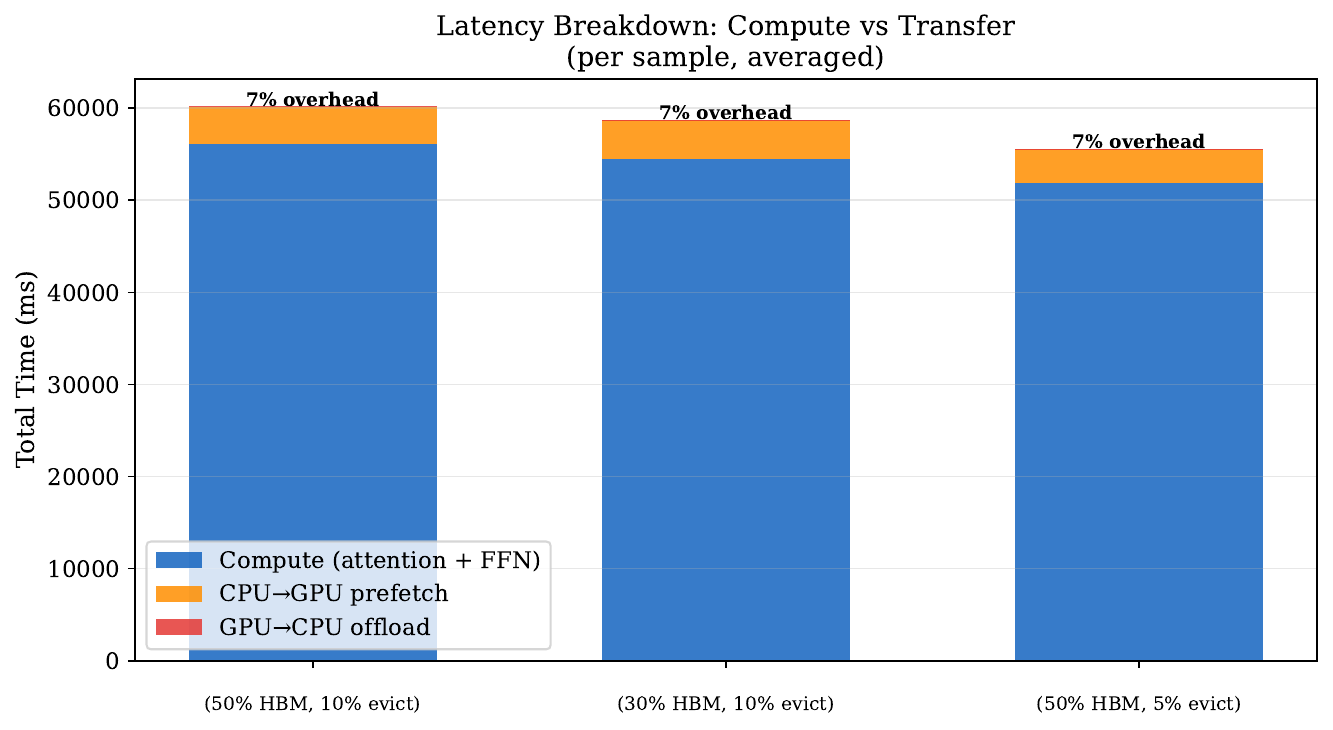}
    \caption{Latency breakdown for hierarchy configurations. Transfer overhead (GPU$\leftrightarrow$CPU offload + prefetch) constitutes only 5--7\% of total inference time; model computation dominates.}
    \label{fig:breakdown}
\end{figure}

\paragraph{Memory analysis and scaling projection.}
In our single-request 7B prototype, the KV cache is small relative to model weights (${\sim}$0.5\,GB vs.\ 14\,GB for fp16, or 7\,GB for int8).
However, KV cache memory grows linearly with both batch size and sequence length, and becomes the dominant memory consumer in production serving scenarios.
Table~\ref{tab:scaling} projects KV cache as a fraction of total GPU memory across representative deployment configurations.

\begin{table}[t]
\centering
\caption{Projected KV cache memory fraction across deployment scenarios. Per-token KV size = $2 \times n_\text{layers} \times n_\text{heads} \times d_\text{head} \times 2$\,bytes (fp16). Our hierarchy offloads 50--70\% of KV cache to DDR, yielding the shown HBM savings.}
\label{tab:scaling}
\vskip 0.1in
\small
\begin{tabular}{llcccc}
\toprule
\textbf{Model} & \textbf{Scenario} & \textbf{KV Cache} & \textbf{Model Wt} & \textbf{KV/Total} & \textbf{Savings} \\
\midrule
7B (fp16) & batch=1, 2K tokens & 0.5\,GB & 14\,GB & 3\% & 0.3\,GB \\
7B (fp16) & batch=8, 2K tokens & 3.8\,GB & 14\,GB & 21\% & 2.3\,GB \\
7B (fp16) & batch=1, 16K tokens & 3.8\,GB & 14\,GB & 21\% & 2.3\,GB \\
7B (int4) & batch=8, 2K tokens & 3.8\,GB & 3.5\,GB & 52\% & 2.3\,GB \\
70B (fp16) & batch=8, 4K tokens & 80\,GB & 140\,GB & 36\% & 48\,GB \\
70B (int4) & batch=8, 4K tokens & 80\,GB & 35\,GB & 70\% & 48\,GB \\
\bottomrule
\end{tabular}
\end{table}

At batch size 8 with a quantized 7B model, KV cache already constitutes 52\% of total GPU memory; for a 70B model with int4 weights and batch serving, KV cache dominates at 70\%.
In these regimes, our hierarchy's 50--70\% KV cache offloading translates to 2--48\,GB of HBM savings---the difference between fitting on a single consumer GPU versus requiring multi-GPU setups.
Our prototype validates the \emph{correctness} and \emph{overhead} of the approach (5--7\% transfer cost, no accuracy loss) at a scale where the absolute memory savings are modest; the architectural benefit compounds as models, sequences, and batch sizes grow.

\subsection{Ablation: Importance Scoring Methods}
\label{sec:scoring_ablation}

We compare four importance scoring strategies, all evaluated with 5\% eviction on GSM8K (Table~\ref{tab:scoring}).
\emph{Attention-only} (our default) computes cumulative attention weight per position.
\emph{VATP}~\cite{vatp} multiplies attention scores by value vector norms, motivated by the finding that attention sinks have high attention but low value norms.
\emph{Redundancy} penalizes tokens with high cosine similarity to neighbors, inspired by R-KV~\cite{rkv}.
\emph{Combined} integrates all three signals.

\begin{table}[t]
\centering
\caption{Importance scoring ablation at 5\% eviction on GSM8K ($n=50$). 95\% CIs in brackets.
Cumulative attention alone outperforms all alternatives.}
\label{tab:scoring}
\vskip 0.1in
\begin{tabular}{lc}
\toprule
\textbf{Scoring Method} & \textbf{GSM8K Accuracy} \\
\midrule
Attention-only (ours) & \textbf{58.0\%} {\scriptsize[43,72]} \\
VATP (attn $\times$ value norm) & 50.0\% {\scriptsize[36,64]} \\
Combined (attn $\times$ val $-$ redundancy) & 46.0\% {\scriptsize[32,61]} \\
Redundancy (attn $-$ redundancy) & 42.0\% {\scriptsize[28,57]} \\
\bottomrule
\end{tabular}
\end{table}

Surprisingly, all alternatives degrade accuracy relative to the attention-only baseline.
The VATP method loses 8 absolute points, and redundancy-aware scoring loses 16 points.
We hypothesize that for reasoning tokens---unlike general text---value norm and neighbor redundancy introduce noise rather than complementary signal.
Reasoning chains contain structurally repetitive but semantically distinct steps (e.g., ``Step 1: ..., Step 2: ...''), causing redundancy penalties to incorrectly suppress important tokens.
The attention-only score, which directly measures each position's contribution to subsequent computation, provides the cleanest importance signal for chain-of-thought reasoning.

\begin{figure}[t]
    \centering
    \includegraphics[width=0.75\textwidth]{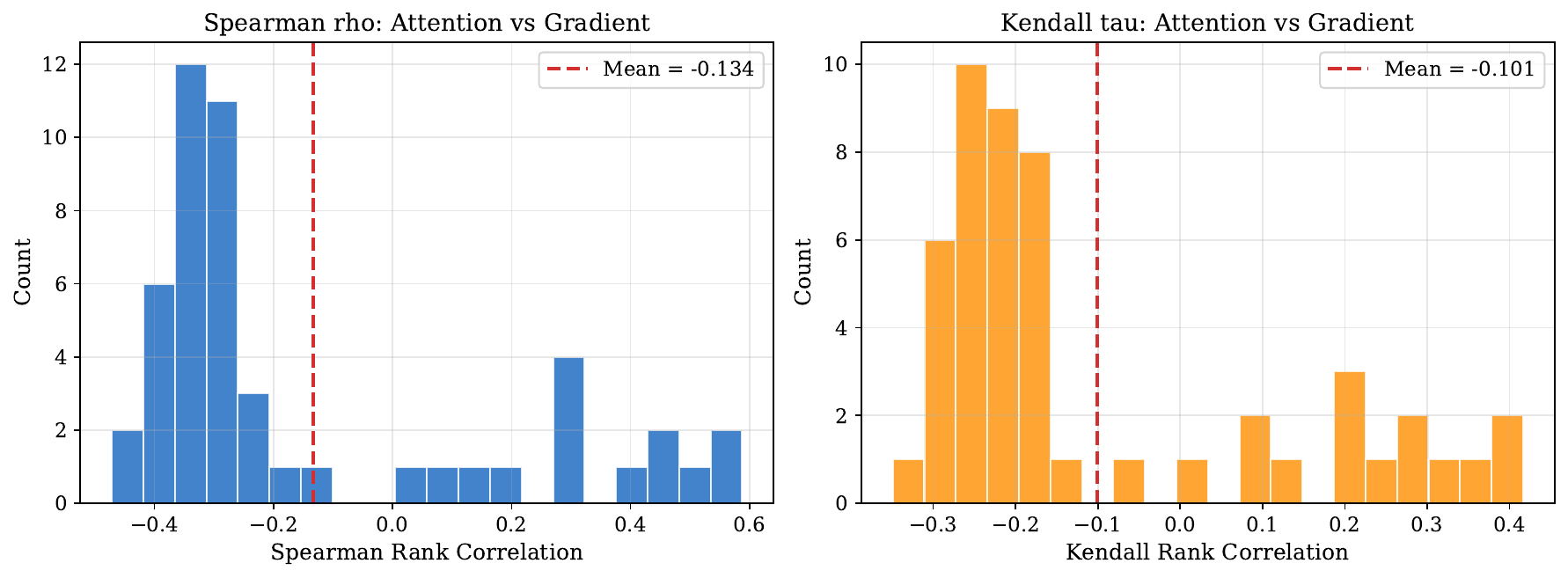}
    \caption{Distribution of rank correlations between attention-based and gradient-based importance scores across 50 GSM8K samples. The near-zero mean Spearman $\rho$ ($-0.13$) indicates the two methods rank tokens in fundamentally different orders.}
    \label{fig:gradient}
\end{figure}

\paragraph{Gradient-based scoring.}
We additionally compare attention-based scoring against gradient-based importance, which computes $\|\partial \mathcal{L}/\partial \mathbf{k}_i\| + \|\partial \mathcal{L}/\partial \mathbf{v}_i\|$ for each position (Table~\ref{tab:gradient}).
The two methods exhibit near-zero rank correlation (Spearman $\rho = -0.13 \pm 0.31$, $n{=}50$; Figure~\ref{fig:gradient}), indicating they capture fundamentally different importance signals.
Despite this, gradient-based scoring achieves comparable streaming eviction accuracy to attention-based scoring at 90\% and 70\% budgets, and slightly worse at 50\% budget (2\% vs.\ 8\%).
We retain cumulative attention as the default because it requires no backward pass (gradient scoring adds ${\sim}$40\% computational overhead) and performs equally or better across all budget levels.

\begin{table}[t]
\centering
\caption{Gradient vs.\ attention scoring for streaming eviction on GSM8K ($n=50$, int8, RTX 5080).}
\label{tab:gradient}
\vskip 0.1in
\begin{tabular}{lcc}
\toprule
\textbf{HBM Budget} & \textbf{Attention} & \textbf{Gradient} \\
\midrule
90\% & 68\% & 70\% \\
70\% & 32\% & 32\% \\
50\% & 8\% & 2\% \\
\bottomrule
\end{tabular}
\end{table}

\section{Related Work}
\label{sec:related}

Existing approaches to KV-cache management span four categories (Table~\ref{tab:landscape}), none of which combines reasoning-aware scoring with cross-tier memory placement.

\paragraph{Eviction.}
StreamingLLM~\cite{streamingllm} maintains attention sinks plus a sliding window; H2O~\cite{h2o} formulates eviction as dynamic submodular optimization; ScissorHands~\cite{scissorhands} exploits persistence of importance.
All permanently discard non-retained tokens.

\paragraph{Compression.}
SnapKV~\cite{snapkv}, KIVI~\cite{kivi}, GEAR~\cite{gear}, MiniCache~\cite{minicache}, and PyramidKV~\cite{pyramidkv} reduce memory via quantization or low-rank decomposition but operate entirely within HBM.

\paragraph{Offloading.}
InfiniGen~\cite{infinigen}, ArkVale~\cite{arkvale}, HeadInfer~\cite{headinfer}, and ScoutAttention~\cite{scoutattention} move KV entries to CPU memory, but their placement policies are reasoning-agnostic.

\paragraph{Reasoning-aware methods.}
ThinKV~\cite{thinkv} classifies CoT segments by type and applies differentiated compression, achieving $<$4\% accuracy drop at 5\% KV retention---but operates entirely within HBM.
R-KV~\cite{rkv} jointly scores by importance and redundancy ($Z = \lambda I - (1{-}\lambda)R$), reporting lossless compression at 10--34\% cache, though its post-RoPE scoring degrades at long sequences~\cite{triattention}.
TriAttention~\cite{triattention} stabilizes scoring via pre-RoPE trigonometric series.
``Hold Onto That Thought''~\cite{holdonto} benchmarks these methods on our same model, confirming that heavy-hitter strategies outperform alternatives but still suffer from eviction's fundamental limitation.

Our work fills the gap in Table~\ref{tab:landscape}: reasoning-aware scoring drives \emph{placement} across tiers, not eviction.

\begin{table}[t]
\centering
\caption{Landscape of KV-cache optimization. Our approach is the first to combine reasoning-aware scoring with tiered offloading.}
\label{tab:landscape}
\vskip 0.1in
\small
\begin{tabular}{lcccc}
\toprule
\textbf{Method} & \textbf{R-Aware} & \textbf{Offload} & \textbf{Quant} & \textbf{Evict} \\
\midrule
H2O~\cite{h2o} & \xmark & \xmark & \xmark & \cmark \\
StreamingLLM~\cite{streamingllm} & \xmark & \xmark & \xmark & \cmark \\
KIVI~\cite{kivi} & \xmark & \xmark & \cmark & \xmark \\
InfiniGen~\cite{infinigen} & \xmark & \cmark & \xmark & \cmark \\
ScoutAttention~\cite{scoutattention} & \xmark & \cmark & \xmark & \cmark \\
ThinKV~\cite{thinkv} & \cmark & \xmark & \cmark & \cmark \\
R-KV~\cite{rkv} & \cmark & \xmark & \xmark & \cmark \\
TriAttention~\cite{triattention} & \cmark & \xmark & \xmark & \cmark \\
\midrule
\textbf{Ours} & \cmark & \cmark & \cmark & \cmark \\
\bottomrule
\end{tabular}
\end{table}

\section{Discussion}
\label{sec:discussion}

\paragraph{Why offloading succeeds where eviction fails.}
The contrast between GSM8K and MATH-500 is instructive.
At 3\% eviction, the hierarchy loses 8.5\% relative accuracy on grade-school math but 29\% on competition problems---a gap that reflects a deeper property of reasoning.
Harder problems pack more information per token: a variable binding or intermediate algebraic identity that seems cold (low recent attention) may become the linchpin of a proof strategy 500 tokens later.
Eviction treats such tokens as expendable; the hierarchy treats them as merely \emph{distant}, storing them in DDR where they can be recalled without approximation.
The zero-error guarantee formalizes what the experiments bear out: the distinction between ``forgotten'' and ``filed away'' is the difference between 0\% and 65\% accuracy.

\paragraph{The information density gradient.}
Our results suggest a useful conceptual tool: different reasoning tasks lie along a spectrum of \emph{per-token information density}.
GSM8K, with its shorter chains and simpler dependencies, tolerates 3\% eviction well (91\% retention).
MATH-500, where chains are denser and errors compound faster, tolerates it less (71\%).
MATH Level-5, with its very long chains approaching 4K tokens, tolerates it least (33\%).
This gradient has practical implications: the ``right'' eviction ratio is not universal but task-dependent, and the hierarchy provides a knob---the eviction ratio---that practitioners can tune without touching HBM allocation.

\paragraph{Limitations and future work.}
Our validation spans three scales (7B--32B) within a single model family (DeepSeek-R1-Distill-Qwen); cross-architecture generalization to Llama or Mistral remains open.
Scaling projections (Table~\ref{tab:scaling}) suggest the hierarchy's HBM savings grow with model size, but empirical validation at 70B+ requires multi-GPU infrastructure.
The T2 (compressed) tier is architecturally defined but practically constrained: 8-bit quantization degrades accuracy from 58\% to 20\%, indicating that reasoning tokens demand precision in ways that current quantization methods do not handle well.
Future work on reasoning-aware compression---perhaps differentiating precision by thought segment~\cite{thinkv}---could unlock this tier.
Finally, our PCIe bandwidth utilization (24--35\% of peak) is limited by unpinned memory; production systems using \texttt{cudaHostAlloc} and batched CUDA streams should reduce overhead below the 5--7\% we report.

\paragraph{A placement layer, not a scoring method.}
We emphasize that the hierarchy is orthogonal to advances in importance scoring.
It does not compete with R-KV's redundancy detection or TriAttention's trigonometric series---it provides the \emph{infrastructure beneath them}.
Any method that currently assigns importance and evicts can instead assign importance and offload.
TriAttention's~\cite{triattention} scoring could replace our cumulative attention for improved long-sequence stability; ThinKV's~\cite{thinkv} segment classification could map directly to tier boundaries.
The hierarchy is the platform; better scoring methods are plugins.

\paragraph{Broader significance.}
The memory hierarchy paradigm becomes more consequential as models grow.
At our 7B prototype scale, HBM savings are modest; at 70B with batch=8, KV cache consumes 70\% of GPU memory, and offloading 48\,GB to DDR could enable single-GPU serving where multi-GPU was previously required (Table~\ref{tab:scaling}).
More broadly, as the field invests in test-time compute scaling---producing ever-longer reasoning chains---the gap between what must be remembered and what must be fast will widen.
Memory hierarchies address this gap with a principle borrowed from six decades of computer architecture: \emph{not all data needs to be equally close to the processor}.

\section{Conclusion}
\label{sec:conclusion}

The dominant paradigm for managing KV-cache memory asks: \emph{which tokens can we afford to throw away?}
For reasoning LLMs, we have shown that this is the wrong question.
Even 3\% permanent eviction costs measurable accuracy; at 50\%, the model collapses entirely.
The right question is: \emph{where should each token live?}

Our semantics-aware memory hierarchy answers this by placing tokens across four tiers---HBM, DDR, compressed, and evicted---according to their real-time importance.
The key theoretical insight is that DDR offloading introduces zero approximation error: prefetched tokens contribute identical attention terms regardless of where they were stored.
This decouples HBM capacity from reasoning quality, yielding an empirical finding confirmed across a $3{\times}3$ grid of configurations, three model scales (7B--32B), and four benchmarks: accuracy depends on the eviction ratio, not the HBM ratio.
With only 3\% eviction, the hierarchy retains 91\% of baseline accuracy on GSM8K ($n{=}200$), and a system prototype confirms that the overhead of cross-tier data movement is just 5--7\% of inference time.

As reasoning chains grow longer with future models---driven by test-time compute scaling---the gap between ``what must be remembered'' and ``what must be fast'' will only widen.
Memory hierarchies offer a fundamentally different path from the eviction-or-nothing paradigm: one where cheaper, more abundant memory absorbs the pressure that HBM cannot sustain alone.

\section*{Reproducibility Statement}

We provide complete implementation details to enable reproduction of our results.
The models (DeepSeek-R1-Distill-Qwen-7B/14B/32B) and datasets (GSM8K, MATH-500, ARC-Challenge) are publicly available on HuggingFace.
All hyperparameters are fully specified: sink size $k_s{=}4$, window size $k_w{=}128$, management interval $\Delta{=}64$, greedy decoding, max 2,048/4,096 new tokens.
Algorithm~\ref{alg:hierarchy} provides complete pseudocode for the tier management loop.
We report 95\% Clopper-Pearson confidence intervals for all accuracy numbers to quantify statistical uncertainty at our sample sizes.
Hardware: RTX 6000 Ada 48\,GB for algorithm experiments (7B/14B), RTX 5080 16\,GB for system prototype, and NVIDIA A100 40\,GB for the HBM independence grid and R-KV baseline reproduction.
Software: PyTorch 2.5, Transformers 4.46, greedy decoding throughout.
Code and experiment scripts will be released upon publication.

\newpage
\appendix
\section{Hyperparameter Sensitivity}
\label{app:sensitivity}

We conduct ablation sweeps on two key hyperparameters of our hierarchy system, evaluated on GSM8K ($n{=}50$) with 50\% HBM ratio and 5\% eviction ratio.

\paragraph{Window size $k_w$.}
Table~\ref{tab:window_sweep} shows accuracy across window sizes $\{32, 64, 128, 256, 512\}$ with fixed manage\_interval$=64$.
Counterintuitively, the smallest window ($k_w{=}32$: 74\%) achieves the highest accuracy, while the default $k_w{=}128$ yields 62\%.
This occurs because the window constitutes a \emph{protected} region exempt from importance-based management: a smaller window frees more token slots for importance-ranked selection, allowing the cumulative attention scoring to place high-value tokens in HBM regardless of recency.
In effect, \textbf{importance scoring is a better allocator than recency}, and shrinking the recency window amplifies its influence.
However, the wide confidence intervals (all overlapping) mean the differences are not statistically significant at $n{=}50$; we retain the conservative $k_w{=}128$ default.

\begin{table}[h]
\centering
\caption{Window size ablation on GSM8K ($n=50$, 7B fp16, 50\% HBM, 5\% eviction). 95\% CIs in brackets.}
\label{tab:window_sweep}
\vskip 0.1in
\begin{tabular}{lc}
\toprule
\textbf{Window Size $k_w$} & \textbf{GSM8K Accuracy} \\
\midrule
32 & 74.0\% {\scriptsize[60,85]} \\
64 & 58.0\% {\scriptsize[43,72]} \\
128 (default) & 62.0\% {\scriptsize[47,75]} \\
256 & 64.0\% {\scriptsize[49,77]} \\
512 & 70.0\% {\scriptsize[55,82]} \\
\bottomrule
\end{tabular}
\end{table}

\paragraph{Management interval $\Delta$.}
Table~\ref{tab:interval_sweep} shows accuracy across intervals $\{16, 32, 64, 128, 256\}$ with fixed window\_size$=128$.
Accuracy is remarkably stable across the full range (60--68\%), confirming that the method is robust to this hyperparameter.
The slight optimum at $\Delta{=}128$ (68\%) is not statistically significant relative to the default $\Delta{=}64$ (62\%; overlapping CIs), and smaller intervals incur proportionally higher sorting overhead.
We retain $\Delta{=}64$ as the default, balancing responsiveness with computational cost.

\begin{table}[h]
\centering
\caption{Management interval ablation on GSM8K ($n=50$, 7B fp16, 50\% HBM, 5\% eviction).}
\label{tab:interval_sweep}
\vskip 0.1in
\begin{tabular}{lc}
\toprule
\textbf{Interval $\Delta$} & \textbf{GSM8K Accuracy} \\
\midrule
16 & 64.0\% {\scriptsize[50,76]} \\
32 & 60.0\% {\scriptsize[46,74]} \\
64 (default) & 62.0\% {\scriptsize[48,76]} \\
128 & \textbf{68.0\%} {\scriptsize[54,80]} \\
256 & 64.0\% {\scriptsize[50,76]} \\
\bottomrule
\end{tabular}
\end{table}

\section{PCIe Transfer Details}
\label{app:pcie}

Figures~\ref{fig:pcie} and~\ref{fig:bandwidth} show detailed PCIe latency and bandwidth measurements on RTX 5080 (PCIe Gen5 x16).
These measurements were conducted by transferring KV cache tensors of varying sizes across all 28 model layers.
The bandwidth asymmetry (22\,GB/s GPU$\to$CPU vs.\ 15\,GB/s CPU$\to$GPU) arises from non-pinned memory allocation; production systems using \texttt{cudaHostAlloc} would improve CPU$\to$GPU throughput.

\section{Gradient-Based Scoring Details}
\label{app:gradient}

Gradient-based importance computes $\|\partial \mathcal{L}/\partial \mathbf{k}_i\| + \|\partial \mathcal{L}/\partial \mathbf{v}_i\|$ using the next-token prediction loss.
This requires a backward pass per step, adding ${\sim}40\%$ computational overhead.
Figure~\ref{fig:gradient} shows that attention and gradient scores exhibit near-zero rank correlation (Spearman $\rho{=}-0.13$), indicating they capture fundamentally different aspects of token importance.
Despite this, both achieve comparable eviction accuracy, suggesting that multiple importance signals converge to similar tier assignments.

\section{Tier Distribution Evolution}
\label{app:tier_evolution}

\begin{figure}[h]
    \centering
    \includegraphics[width=0.7\textwidth]{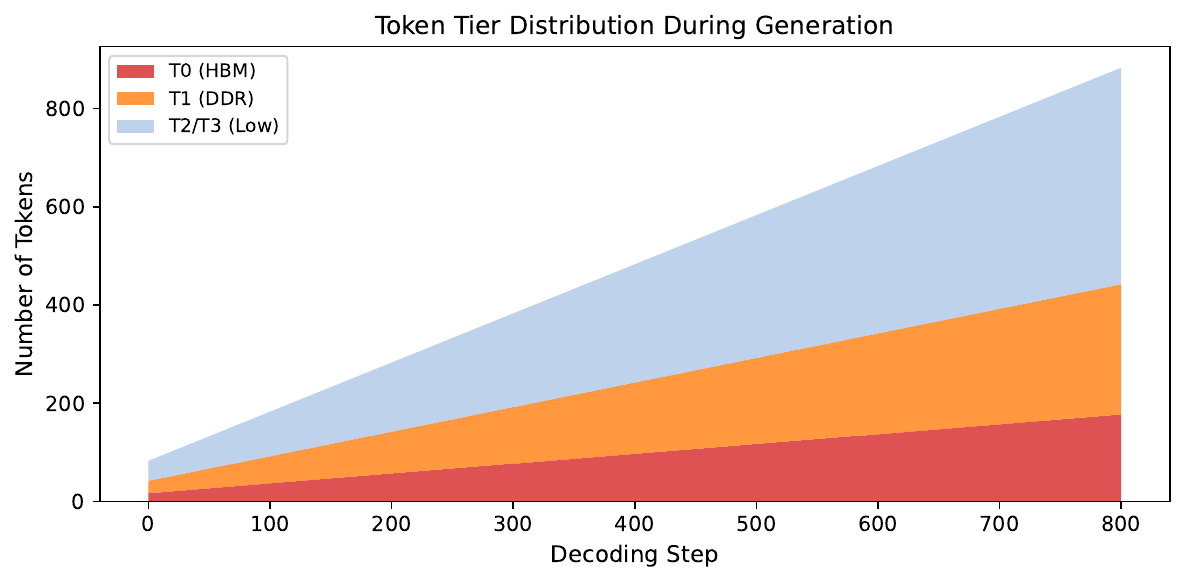}
    \caption{Token tier distribution during generation. The hierarchy dynamically assigns tokens to tiers as the chain grows, with T0 (HBM) capacity remaining bounded.}
    \label{fig:tier_evo}
\end{figure}

Figure~\ref{fig:tier_evo} shows how the number of tokens in each tier changes during generation.
As the reasoning chain grows, the proportion of T0 (HBM) tokens remains stable while T1 (DDR) and T2/T3 tokens grow proportionally, demonstrating that the hierarchy naturally adapts to sequence length.

\section{R-KV Baseline Reproduction}
\label{app:rkv}

To ensure a fair comparison with R-KV~\cite{rkv}, we re-implemented its core algorithm on our setup rather than comparing across papers.
R-KV scores each token using a joint importance-redundancy metric:
\begin{equation}
    Z_i = \lambda \cdot I_i - (1-\lambda) \cdot R_i
\end{equation}
where $I_i$ is the attention-based importance (softmax of the last $\alpha{=}8$ observation tokens, max-pooled with kernel size 7), $R_i$ is the cosine-similarity-based redundancy score, and $\lambda{=}0.07$ balances the two signals.
Tokens with the lowest $Z_i$ are evicted to maintain a fixed-size KV cache budget.

We evaluated at three budget levels on the same 50 GSM8K samples used in our hierarchy experiments (Table~\ref{tab:rkv_detail}).
The DynamicCache compatibility was ensured by converting between tuple and DynamicCache formats as required by Transformers 4.46.

\begin{table}[h]
\centering
\caption{R-KV reproduction results on GSM8K ($n{=}50$, DeepSeek-R1-Distill-Qwen-7B, A100 40GB). All runs use greedy decoding, max 2048 tokens.}
\label{tab:rkv_detail}
\vskip 0.1in
\begin{tabular}{lccc}
\toprule
\textbf{Budget} & \textbf{Correct} & \textbf{Accuracy} & \textbf{vs.\ Hierarchy 3\%} \\
\midrule
256 tokens & 0/50 & 0.0\% & 0$\times$ \\
512 tokens & 3/50 & 6.0\% & 0.10$\times$ \\
1024 tokens & 16/50 & 32.0\% & 0.53$\times$ \\
\midrule
\multicolumn{2}{l}{Hierarchy (3\% evict)} & 58--62\% & 1.0$\times$ \\
\bottomrule
\end{tabular}
\end{table}

At budget=256, R-KV generates incoherent text---the aggressive compression destroys the reasoning chain entirely.
At budget=1024, accuracy recovers to 32\%, but this is still roughly half of what the hierarchy achieves with 3\% eviction.
The gap is not explained by scoring quality: R-KV's importance+redundancy scoring is more sophisticated than our cumulative attention.
Rather, it reflects the fundamental limitation of eviction for reasoning: discarding tokens, regardless of how carefully they are selected, destroys information that offloading preserves.

\section{Additional Experimental Details}
\label{app:details}

\paragraph{Answer extraction.}
We extract numeric answers from model outputs using a cascading pattern matcher:
(1)~\texttt{\#\#\#\#} delimiter (GSM8K convention),
(2)~``the answer is X'' patterns,
(3)~\texttt{\textbackslash boxed\{X\}} LaTeX format,
(4)~last standalone number in the output.
Answers are compared numerically with tolerance $10^{-5}$.

\paragraph{Confidence intervals.}
All accuracy numbers report 95\% Clopper-Pearson (exact binomial) confidence intervals.
At $n{=}50$, these intervals are $\pm$12--15\%; at $n{=}200$, they narrow to $\pm$6--7\%.
We use Fisher's exact test for pairwise significance rather than asymptotic tests, as several conditions have near-zero accuracy.

\paragraph{Hyperparameters.}
All hierarchy experiments use: sink size $k_s{=}4$, window size $k_w{=}128$, management interval $\Delta{=}64$, HBM ratio $\beta{=}0.5$ (unless varied in the grid), eviction ratio $r$ as specified per experiment.
Greedy decoding with max 2,048 new tokens (4,096 for MATH Level-5).
No prompt engineering; questions are presented in a minimal ``solve step by step'' format.

\section{End-to-End System Visualization}
\label{app:e2e_vis}

\begin{figure}[h]
    \centering
    \includegraphics[width=\textwidth]{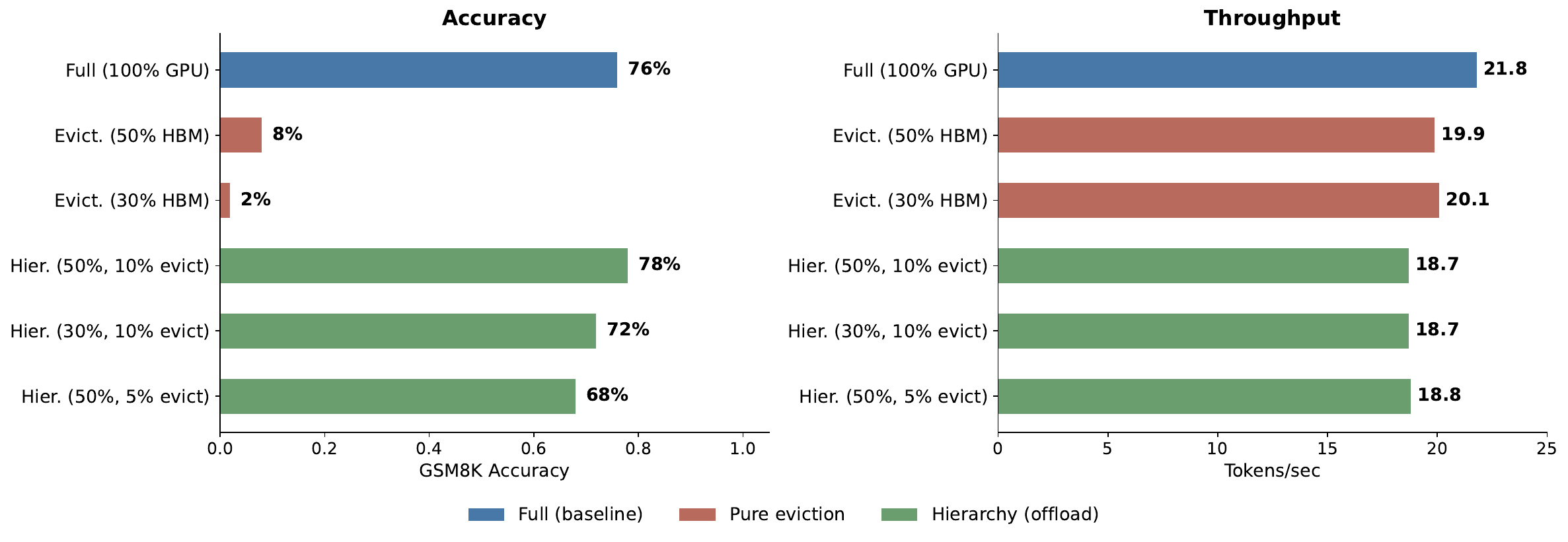}
    \caption{End-to-end accuracy and throughput comparison. Hierarchy configurations (green) achieve comparable accuracy to the full baseline (blue) with modest throughput reduction. Eviction (red) collapses in accuracy.}
    \label{fig:e2e_vis}
\end{figure}

Figure~\ref{fig:e2e_vis} provides a visual summary of the system prototype results: hierarchy configurations cluster near full-cache accuracy while maintaining $>$85\% throughput.
Figure~\ref{fig:breakdown} breaks down the latency into compute and transfer components.

\begin{figure}[h]
    \centering
    \includegraphics[width=0.7\textwidth]{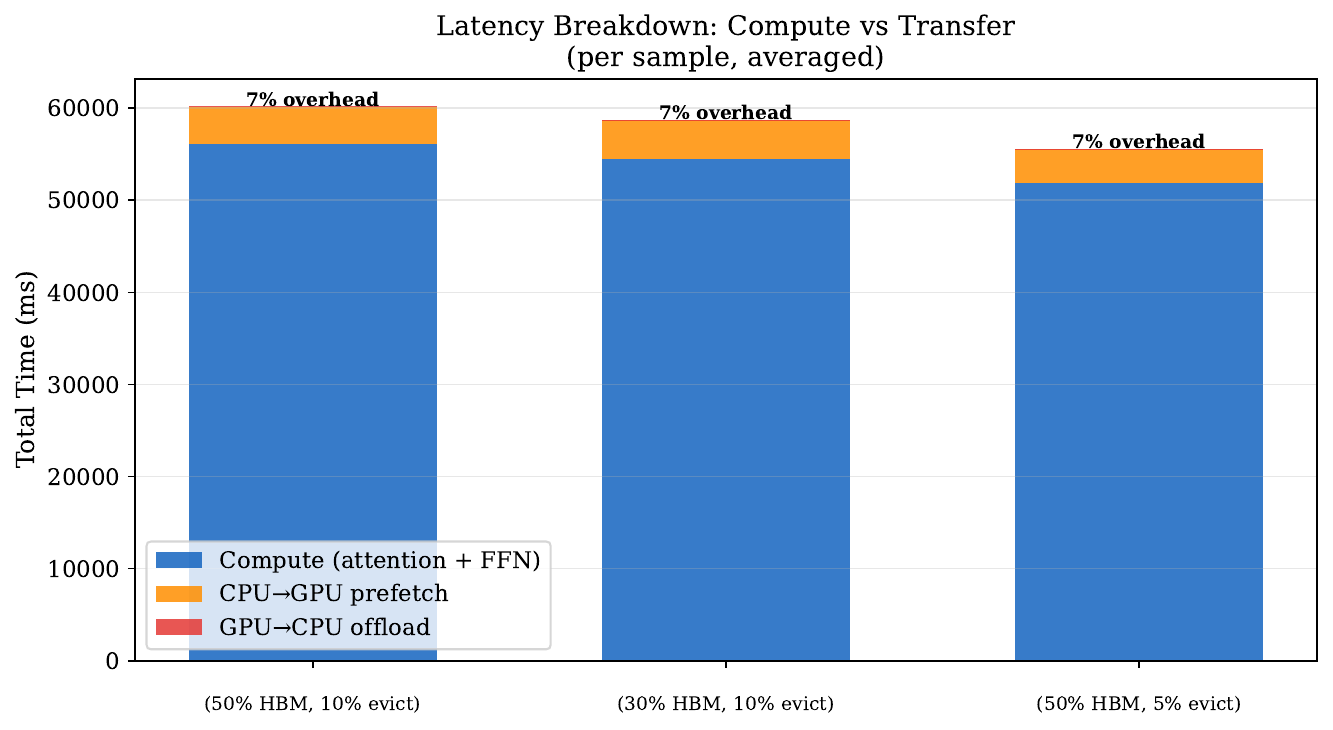}
    \caption{Latency breakdown per sample (averaged). Transfer overhead (GPU$\leftrightarrow$CPU) constitutes only 5--7\% of total inference time.}
    \label{fig:breakdown}
\end{figure}

\section{Scaling Projections}
\label{app:scaling}

Table~\ref{tab:scaling} projects KV cache as a fraction of total GPU memory across deployment scenarios.
At batch=8 with int4 quantization, KV cache already dominates at 52\%; for 70B models it reaches 70\%.

\begin{table}[h]
\centering
\caption{Projected KV cache memory fraction. Our hierarchy offloads 50--70\% to DDR.}
\label{tab:scaling}
\vskip 0.1in
\small
\begin{tabular}{llcccc}
\toprule
\textbf{Model} & \textbf{Scenario} & \textbf{KV} & \textbf{Weights} & \textbf{KV/Total} & \textbf{Savings} \\
\midrule
7B fp16 & bs=1, 2K tok & 0.5\,GB & 14\,GB & 3\% & 0.3\,GB \\
7B fp16 & bs=8, 2K tok & 3.8\,GB & 14\,GB & 21\% & 2.3\,GB \\
7B int4 & bs=8, 2K tok & 3.8\,GB & 3.5\,GB & 52\% & 2.3\,GB \\
70B fp16 & bs=8, 4K tok & 80\,GB & 140\,GB & 36\% & 48\,GB \\
70B int4 & bs=8, 4K tok & 80\,GB & 35\,GB & 70\% & 48\,GB \\
\bottomrule
\end{tabular}
\end{table}


\end{document}